\documentclass{article}

\newcommand{\nameshort}{VaLSe}

    \usepackage[preprint]{neurips_2025}

\usepackage[utf8]{inputenc} 
\usepackage[T1]{fontenc}    
\usepackage{hyperref}       
\usepackage{url}            
\usepackage{booktabs}       
\usepackage{amsfonts}       
\usepackage{nicefrac}       
\usepackage{microtype}      
\usepackage{xcolor}         
\usepackage{amsmath}
\usepackage{graphicx}
\usepackage{amssymb}
\usepackage{multirow}

\usepackage{colortbl}
\definecolor{mygray}{gray}{.92}
\usepackage{caption}

\usepackage{enumitem}

\renewcommand{\tilde}{\widetilde}

\def\eqref#1{equation~\ref{#1}}

\def\1{\bm{1}}

\def\rvh{{\mathbf{h}}}

\def\rmA{{\mathbf{A}}}

\def\rmC{{\mathbf{C}}}

\def\rmM{{\mathbf{M}}}

\def\rmV{{\mathbf{V}}}
\def\rmW{{\mathbf{W}}}

\def\vx{{\bm{x}}}

\def\vx{{\boldsymbol{x}}}

\DeclareMathAlphabet{\mathsfit}{\encodingdefault}{\sfdefault}{m}{sl}
\SetMathAlphabet{\mathsfit}{bold}{\encodingdefault}{\sfdefault}{bx}{n}

\title{Seeing It or Not? Interpretable Vision-aware Latent Steering to Mitigate Object Hallucinations}

\author{%
  Boxu Chen\thanks{Equal contribution.}\hspace{2mm}, Ziwei Zheng$^{*}$, Le Yang\thanks{Corresponding Author.}\hspace{2mm}, Zeyu Geng, \\ \textbf{Zhengyu Zhao},\textbf{Chenhao Lin}, \textbf{Chao Shen}
  \\ 
  Xi'an Jiaotong University \\ 
  \texttt{\{chenboxu, ziwei.zheng, gengzeyu\}@stu.xjtu.edu.cn}, \\ 
  \texttt{\{yangle15, zhengyu.zhao, chenhao.lin, chaoshen\}@xjtu.edu.cn}
}

\begin{document}

\maketitle

\begin{abstract}
Large Vision-Language Models (LVLMs) have achieved remarkable success but continue to struggle with object hallucination (OH), generating outputs inconsistent with visual inputs. While previous work has proposed methods to reduce OH, the visual decision-making mechanisms that lead to hallucinations remain poorly understood. In this paper, we propose VaLSe, a Vision-aware Latent Steering framework that adopts an interpretation-then-mitigation strategy to address OH in LVLMs. By tackling dual challenges of modeling complex vision-language interactions and eliminating spurious activation artifacts, VaLSe can generate visual contribution maps that trace how specific visual inputs influence individual output tokens. These maps reveal the model’s vision-aware focus regions, which are then used to perform latent space steering, realigning internal representations toward semantically relevant content and reducing hallucinated outputs. Extensive experiments demonstrate that VaLSe is a powerful interpretability tool and an effective method for enhancing model robustness against OH across multiple benchmarks. Furthermore, our analysis uncovers limitations in existing OH evaluation metrics, underscoring the need for more nuanced, interpretable, and visually grounded OH benchmarks in future work. Code is available at: \url{https://github.com/Ziwei-Zheng/VaLSe}.

\end{abstract}

\section{Introduction}

Recent advances in large language models (LLMs)~\cite{bai2023qwenllm,touvron2023llama,touvron2023llama2} have accelerated the development of Large Vision-Language Models (LVLMs), such as LLaVA~\cite{liu2024visual,liu2023improved}, InstructBLIP~\cite{dai2023Instructblip}, MiniGPT-4~\cite{zhu2023minigpt4}, and Qwen2-VL~\cite{bai2023qwen,wang2024qwen2}. However, LVLMs are prone to object hallucination~\cite{bai2024hallucination,yang2024nullu,duan2025truthprint,zhou2023analyzing}, often generating outputs that are inconsistent with visual inputs, which raises serious concerns about the reliability and safety of LVLMs. Recent efforts to mitigate hallucinations have explored a range of strategies, including end-to-end fine-tuning~\cite{liu2023mitigating,jiang2024hallucination,kim2023exposing}, post-processing of model outputs~\cite{leng2024mitigating,zhang2024debiasing,zhou2023analyzing,chen2024halc}, and latent feature steering~\cite{yang2024nullu,chen2024ict,liu2025reducing}, all of which have shown promising results on open-source LVLMs. Nevertheless, a critical limitation remains~\cite{bai2024hallucination}: there still lacks an effective method to trace how visual inputs influence the decision-making processes of LVLMs. As a result, the underlying mechanisms of hallucination and the factors triggering it remain poorly understood.

\begin{figure}[t!]
  \centering
  \includegraphics[width=\linewidth]{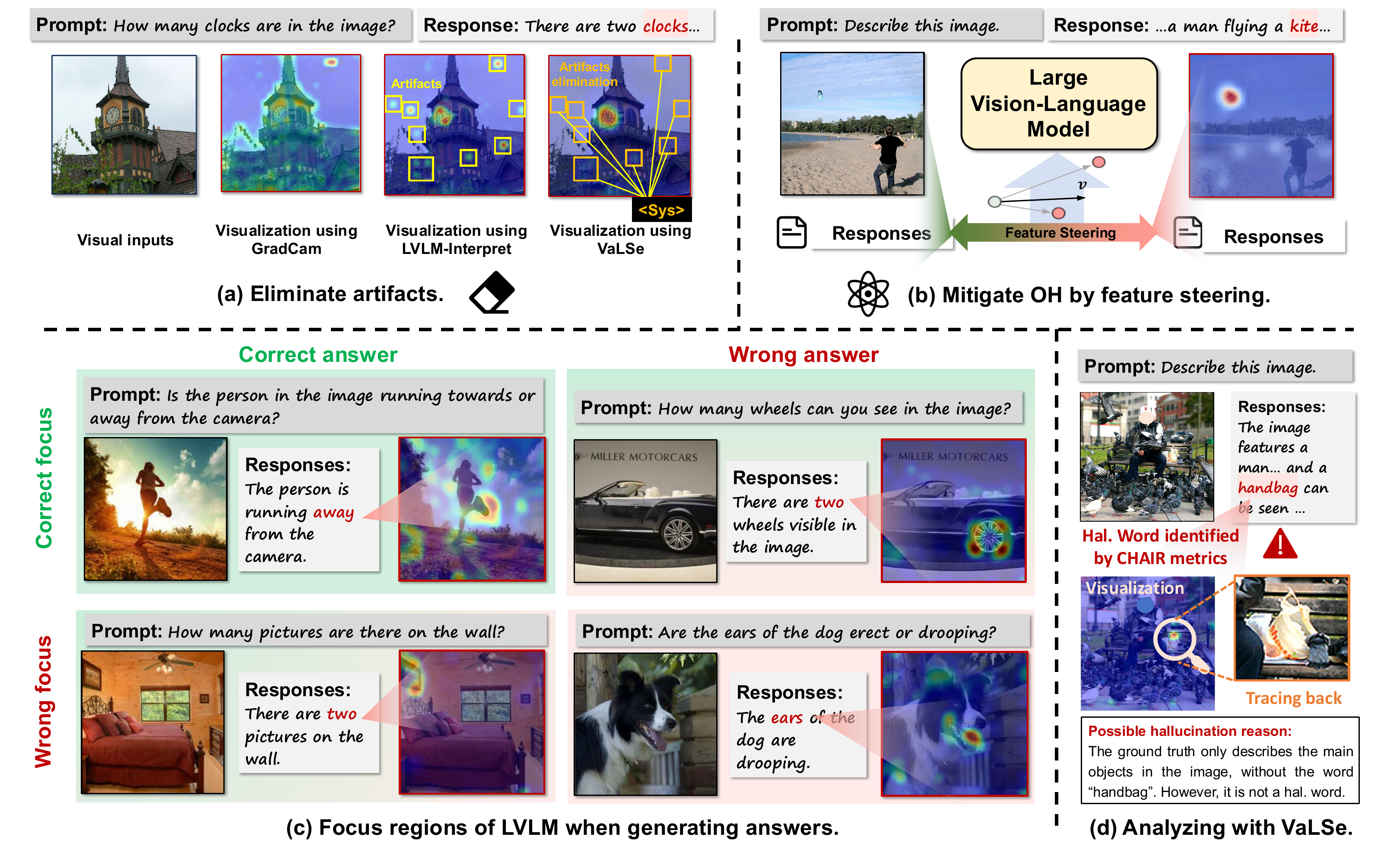}
  \vspace{-15pt}
  \caption{The proposed \nameshort{} can effectively (a) eliminate artifacts and provide high quality visualization results, and then (b) mitigate OH by vision-aware latent steering. With the ability of mitigating OH, \nameshort{} can further provide in-depth analysis of (c) how a word token is generated based on visual information and (d) inferring why a hallucinated words is generated.}
  \label{fig1}
\vspace{-20pt}
\end{figure}

Interpreting open-ended responses from LVLMs introduces several key challenges. (1) Complex vision-language interaction: The intricate alignment between vision encoders and LLMs creates difficulty in disentangling modality contributions and leads to poor interpretable results~\cite{xing2025large,stan2024lvlmintrepret}. (2) Activation artifacts: Recent studies~\cite{kang2025see,darcet2023vision,sun2024massive} reveal that some neurons produce disproportionately high activations regardless of the input, which distort visualization results (see Figure~\ref{fig1}(a)). These challenges hinder the development of reliable interpretation methods for LVLMs, making it difficult to explore why a hallucinated word is generated and to determine whether a correct prediction is a correct answer or a ``guessing one''.

To address these issues, we propose \nameshort{}, a novel \textbf{V}ision-\textbf{a}ware \textbf{L}atent \textbf{S}t\textbf{e}ering framework for LVLMs. Figure~\ref{fig1} provides an overview of \nameshort{}. To trace the influence of visual inputs on output tokens, \nameshort{} models complex vision-language interactions via visual contribution maps and eliminates artifact activations by contrasting targeted tokens with non-semantic special tokens, resulting in higher-quality, interpretable visualization of LVLM's output. Moreover, with the interpretable results, \nameshort{} can reinforce the LVLM’s attention to semantically relevant regions in the image by latent feature steering using the generated visual contribution maps, enhancing its visual grounding and effectively mitigating OH (shown in Figure~\ref{fig1} (b)).

Through comprehensive experiments, we demonstrate that \nameshort{} effectively mitigates OH without compromising general ability. More importantly, \nameshort{} offers a new perspective for studying OH by providing fine-grained interpretability into the model’s decision-making process. As illustrated in the visualization results in Figure~\ref{fig1}, which highlight the focus regions of LLaVA-1.5 during response generation, benchmark ground-truth answers alone are insufficient for determining whether hallucination has occurred. On one hand, a model may produce a correct response while attending to irrelevant image regions, indicating it relied on language priors rather than visual cues. For example, in the bottom-left panel of Figure~\ref{fig1}~(c), the model correctly predicts the word ``two'' without attending to the correct relevant visual evidence. On the other hand, as revealed through visualization (Figure~\ref{fig1}~(d)), a word flagged as hallucinated by metrics (e.g., the CHAIR metric~\cite{rohrbach2018object}) may actually be a visually grounded and accurate description. These findings highlight not only the importance of understanding the internal mechanisms behind hallucinated outputs but also the need for more sophisticated and comprehensive benchmarks to evaluate OH in LVLMs. 

The main contributions are summarized as follows:
\vspace{-3pt}
\begin{itemize}[left=0pt]
    \item We propose a novel vision-aware latent steering method that follows an interpretation-then-mitigation strategy, enabling internal analysis of the generation process behind hallucinated words and effectively reducing OH in LVLMs.
    \item \nameshort{} generates high-quality visual contribution maps across different LVLMs, enabling deeper analysis of their decision-making processes. Our analysis reveals limitations in existing OH evaluation metrics, highlighting the need for more nuanced visually grounded assessment methods.
    \item Experiments demonstrate the effectiveness of \nameshort{} in OH mitigation. Moreover, both qualitative and quantitative evaluations demonstrate the superiority of our method in visualization for LVLMs. 
\end{itemize}

\section{Related Work}

\paragraph{Large Visual-Language Models (LVLMs)}

Based on successes of LLMs, large vision-language models (LVLMs) have made significant progress in recent years. These models typically integrate a vision encoder with an LLM via fusion modules, such as a linear projection layer~\cite{liu2024visual} or a Q-former~\cite{zhu2023minigpt4}. Recent LVLMs, such as LLaVA~\cite{liu2024visual,liu2023improved}, MiniGPT-4~\cite{zhu2023minigpt4}, mPLUG-Owl~\cite{ye2024mplug,ye2024mplugowl3longimagesequenceunderstanding}, Qwen-VL~\cite{bai2023qwenllm,bai2023qwen}, LLaVA-Phi~\cite{zhu2024llava} and DeepSeek-VL~\cite{lu2024deepseek} have shown to be capable for complex image understanding and reasoning. Despite these advancements, modern LVLMs continue to face significant security and robustness challenges, notably object hallucination~\cite{bai2024hallucination}.

\paragraph{Mitigation of Object Hallucination}
Various approaches have been proposed to address this issue. Given that hallucinations may stem from data biases and the knowledge gap between visual and linguistic information, recent studies have explored fine-tuning LVLMs for robustness~\cite{liu2023mitigating,gunjal2024detecting}, cross-modality matching~\cite{jiang2024hallucination,kim2023exposing}, and preference alignment~\cite{sun2023aligning,chen2024dress}.

To avoid the high cost of fine-tuning, post-processing strategies have been developed to revise model outputs using external tools, such as LURE~\cite{zhou2023analyzing} and visual-guided refiners~\cite{yin2023woodpecker,zhao2024mitigating,chen2024halc}. Other approaches aim to debias strong language priors during decoding~\cite{leng2024mitigating,liu2024paying,zhang2024debiasing,zhu2024ibd,huang2024opera,favero2024multi}, while feature-steering methods~\cite{yang2024nullu,liu2025reducing,fang2024alphaedit} learn latent shift directions to adjust internal features for OH mitigation. In contrast, \nameshort{} not only mitigates OH but also interprets the LVLM’s internal generation process, providing insight into the root causes of hallucination. Although ALGA~\cite{an2024agla} also leverages Grad-CAM to generate saliency-based prompts, it relies on an external multimodal model, making it incapable of explaining the LVLM’s own decision-making. \nameshort{}, by contrast, operates entirely within the LVLM and utilizes its interpretability to directly and effectively reduce OH.

\paragraph{Interpretation of LVLM.}
Interpreting computer vision algorithms often involves generating heatmaps that highlight the relevance of different image regions to the model’s decisions. Classical approaches such as Grad-CAM~\cite{selvaraju2017grad} and Grad-CAM++\cite{chattopadhay2018grad} achieve this by combining input feature maps with class-specific gradients from the upper layers of convolutional networks. More recently, transformer interpretability has gained growing attention~\cite{chefer2021generic,chefer2021transformer,aflalo2022vl}, motivating deeper insights into model behavior for interpreting modern LVLMs~\cite{stan2024lvlmintrepret,xing2025large,stan2024fastrm,giulivi2024explaining,zhang2024redundancy,pan2023finding}. In contrast to these interpretability techniques~\cite{stan2024lvlmintrepret,xing2025large}, our method not only provides clearer visual explanations but also leverages them in a knowledge editing framework, leading to more accurate and reliable outputs by mitigating object hallucinations.


\begin{figure}
  \centering
  \includegraphics[width=\linewidth]{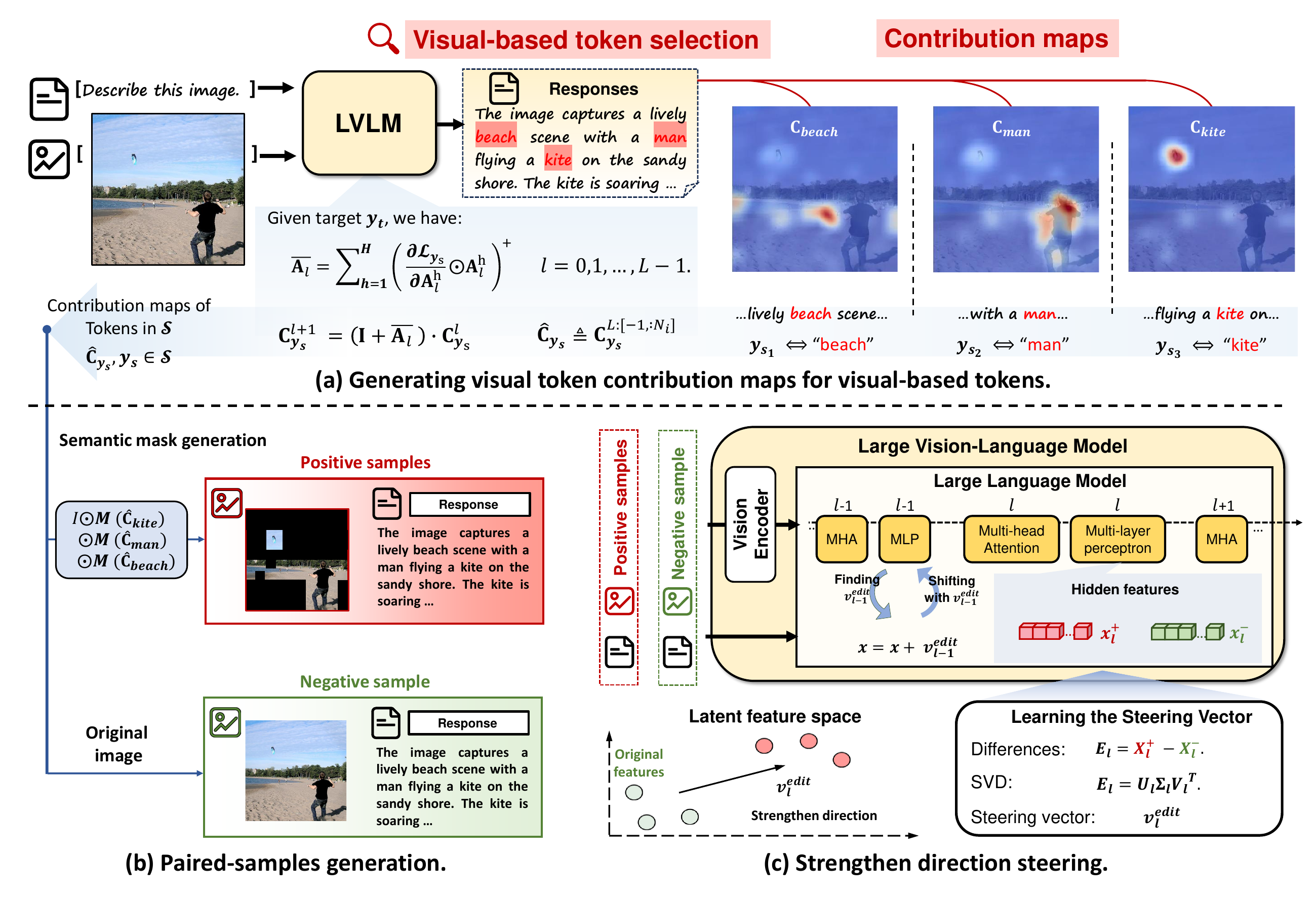}
  \vspace{-15pt}
  \caption{\nameshort{} mainly contains three modules: (a) A visualization module that generates visual token contribution maps for each selected token; (b) A vision-aware masking module creating masked images while preserving the main semantic contents; (c) A latent steering mechanism.}
  \vspace{-15pt}
  \label{fig2}
\end{figure}

\section{Method}
\label{sec:method}

We first present the preliminaries for our method, then introduce the main components of the proposed \nameshort{} and, finally, provide a brief discussion of \nameshort{}. 

\subsection{Preliminaries and Notations} 
Suppose we have an LVLM consisting of an image encoder, an alignment module and an LLM with $L$ layers. In the LLM, the hidden states $\rvh_l$ at layer $l$ can then be calculated as
\begin{align}
\label{eq-1}
\boldsymbol{h}_l = \boldsymbol{x}_l + \boldsymbol{a}_l, ~\text{where} ~~\boldsymbol{x}_l =  \rmW^{\text{out}}_{l}~\sigma\left(\rmW^{\text{in}}_{l} \left(\boldsymbol{a}_l + \boldsymbol{h}_{l-1}\right)\right)~,~ \boldsymbol{a}_l = \sum_{h=1}^H Q_{l}^h (\mathbf{A}_l^h \rmV_l^h).
\end{align}
Here, $\boldsymbol{a}_l$ and $\boldsymbol{x}_l$ represent the outputs of the multi-head attention (MHA) and the multi-layer perceptron (MLP), respectively. The MLP consists of two linear layers with weights $\rmW^{\text{in}}_l$ and $\rmW^{\text{out}}_l$, and an activation function $\sigma$. The attention output $\boldsymbol{a}_l$ is computed by aggregating $H$ attention heads. Each head applies an attention map $\mathbf{A}_l^h$ to its corresponding value matrix $\rmV_l^h$, followed by a projection using $Q_l^h$. For simplicity, layer normalization is omitted from Eq.~\ref{eq-1}.

During autoregressive text generation, words are tokenized and sequentially predicted conditioned on previous tokens. Suppose the answer $y$ consists of $N_r$ tokens, represented as a sequence $y=\left[y_1, y_2\cdots y_{N_r}\right]$. At each step $t$, the model samples the next token $y_t$ according to:
\begin{align}
    y_t\sim P(y_t|y_1, y_2\cdots y_{t-1};I,T),
\end{align}
where $I$ and $T$ are the input image and text, respectively. 


\subsection{\nameshort{}}

\paragraph{Overview.} Figure~\ref{fig2} illustrates the main components of \nameshort{}: (a) Visual-based token selection and contribution map generation, (b) Steering sample construction, and (c) Vision-aware latent steering.

The procedure can be: Given an input image $I$ and text prompt $T$, the LVLM first generates a response $y$. \nameshort{} then selects visual-based tokens whose predictions are strongly influenced by visual input. For each selected token, \nameshort{} computes a visual token contribution map, highlighting the image regions the model attends to during token prediction. These maps are then used to construct positive and negative samples for latent steering. The original image and response serve as the negative sample, while positive ones are created by masking obscure, visually irrelevant regions while preserving core vision-aware objects. Finally, \nameshort{} performs latent steering by computing the directional difference between positive and negative features, adjusting internal representations to reinforce focus on semantically relevant objects and reduce OH.

\paragraph{Visual-based Token Selection.} A visual-based token is defined as one whose prediction confidence is highly sensitive to the presence of visual information. Following~\cite{xing2025large,favero2024multi}, we use the log-likelihood ratio (LLR) between the token’s prediction with and without visual context.

Given $I$, $T$, and the generated responses $y_{<t}$, the probability of token $y_t$ is $P(y_t|y_{<t}, I, T)$. To isolate the influence of the image, we can replace $I$ with a noise image $\Tilde{I}$ that lacks useful visual information, and compute the probability $P(y_t|y_{<t}, \Tilde{I}, T)$. This can be obtained through a single forward pass by concatenating $T$ and $y_{<t}$ as the textual input. The LLR for token $y_t$ is then defined as:
\begin{align}
\label{eq:llr}
\text{LLR}(y_t) = \log P(y_t|y_{<t}, I, T) - \log P(y_t|y_{<t}, \Tilde{I}, T).
\end{align}

A higher value of $\text{LLR}(y_t)$ represents that the token $y_t$ is generated more highly based on visual inputs. We select tokens with high LLR values—those most influenced by the image. Specifically, we define the set of $S$ visual-sensitive tokens as:
\begin{align}
\mathcal{S} = \{ y_s \mid \text{LLR}(y_s) > \alpha,~ s \ne 1 \},
\end{align}
where $\alpha$ is a predefined threshold and $|\mathcal{S}|=S$. The resulting token set $\mathcal{S}$ represents the word tokens in the generated response that are strongly grounded in visual content, which is suitable for visualization\footnote{Note that the proposed \nameshort{} can be used to visualize any token in the response.}. For each of the selected visual-based tokens, we compute the corresponding visual token contribution maps to analyze how the image influences the model’s predictions.

\paragraph{Visual Token Contribution Maps.}

Following~\cite{chefer2021generic}, we compute contribution maps that estimate the relevance of each image token to a specific text token, using the attention mechanisms within the LLM. Let $N_t$\footnote{$N_t$ includes both the original text prompt tokens and the generated responses.} and $N_i$ denote the number of text and image tokens, respectively. The attention map at layer $l$ is represented as $\mathbf{A}_l \in \mathbb{R}^{(N_i + N_t) \times (N_i + N_t)}$.

We then generate the visual contribution map $\mathbf{C}_{y_s}$ for $y_s$, which can be initialized as an identity matrix and propagated layer-by-layer using the attention matrix $\rmA_l$. Since each attention layer has $H$ heads, we follow~\cite{chefer2021transformer} and compute a weighted average of the heads using their gradients with respect to $y_s$. The aggregated attention map $\bar{\mathbf{A}}_l$ at layer $l$ and propagation of $\mathbf{C}_{y_s}$ can be represented as:
\begin{align}
\label{eq:modified_att}
\mathbf{\bar{A}}_l = \sum_{h=1}^H \left( \frac{\partial \mathcal{L}_{y_s}}{\partial \mathbf{A}^h_l} \odot \mathbf{A}^h_l \right)^+, ~~~ \mathbf{C}^{l+1} = \mathbf{C}^l + \mathbf{\bar{A}}_l \cdot \mathbf{C}^l, ~~~l= 0,1,...L-1,
\end{align}
where $\odot$ denotes the element-wise product and $(\cdot)^+$ indicates removing negative contributions.

This iterative update propagates relevance scores from the 0-th layer to the $ L$-th layer. Since the model typically predicts words based on the last token’s hidden state, we take the last row of $\mathbf{C}^L$ and retain the first $N_i$ values, corresponding to the image tokens, $\hat{\mathbf{C}}_{y_s} \triangleq \mathbf{C}_{y_s}^{L [-1,:N_i]}$. Reshaping $\hat{\mathbf{C}}_{y_s}$ yields the visual contribution map for token $y_s$.

\paragraph{Artifacts Elimination.} Generally, $\hat{\mathbf{C}}_{y_s}$ can be significantly affected by artifact activations, which are neurons that consistently exhibit abnormally high values regardless of the input. These artifacts distort the accurate contribution distribution and compromise interpretability.

Following the observation in~\cite{sun2024massive} that such activations typically occur at fixed spatial positions, we address this issue by contrasting contribution maps between target visual-based tokens and a non-semantic special token $y_{sys}$. Specifically, for $y_{sys}$, we compute its contribution map $\hat{\rmC}_{sys}$ and identify positions $\mathcal{P}$ exhibiting artifacts. By suppressing these regions in $\hat{\rmC}_{y_s}$, we obtain cleaner and more accurate visualizations, better reflecting the model’s true attention to image content. We provide a more detailed artifacts elimination procedure in the supplementary materials.

\paragraph{Paired-sample Generation.} For $N$ samples, we first select $N_s$ vision-aware ones, whose $\mathcal{S}$ is not empty, and mask while preserving key visual information indicated by the selected visual-based tokens for each sample. For the $n$-th vision-aware sample, we mask $p$\% image patches associated with low contribution values based on the visual contribution maps, ensuring that important visual content remains intact. Specifically, we will have $S^n$ masks for the $n$-th sample, which can be denoted as $\mathcal{M}_n = \{\rmM(\hat{\mathbf{C}}_{y_s^n}, p) \mid y_s \in \mathcal{S}^n \}$. Applying $\mathcal{M}_n$ yields the masked images ${\tilde{I}_{n}} = I_n \odot \rmM(\hat{\mathbf{C}}_{y_1}, p) \odot  ...\rmM(\hat{\mathbf{C}}_{y_{S^n}}, p) $ for the $n$-th sample, which are then combined with the original text responses, $y$, to form the positive sample. The original image $I_n$ and $y$ can constitute the negative sample. Finally, we have $N_s$ negative and positive samples, all of which will be used to perform vision-aware latent steering.


\paragraph{Vision-aware Latent Steering.} At each layer $l$ of the LLM within the LVLM, we extract features at the MLP for both positive and negative samples. For the $n$-th sample, let $\vx_l^+$ denote the features for the positive samples, and $\vx_l^-$ denote the negative features, we stack them into matrices $\boldsymbol{X}_l^+$ and $\boldsymbol{X}_l^-$, with shape $\mathbb{R}^{N_s \times D}$, where $D$ is the hidden dimension. We then compute the difference matrix $\boldsymbol{E}_l$ at layer $l$ as:
\begin{align}
\boldsymbol{E}_l = \boldsymbol{X}_l^+ - \boldsymbol{X}_l^-, \quad \boldsymbol{E}_l \in \mathbb{R}^{N_s \times D}.
\end{align}

We apply singular value decomposition (SVD) to $\boldsymbol{E}_l$:
\begin{align}
\boldsymbol{E}_l = \boldsymbol{U}_l \boldsymbol{\Sigma}_l \boldsymbol{V}_l^\top, \quad \boldsymbol{U}_l \in \mathbb{R}^{N_s \times N_s}, \quad \boldsymbol{V}_l \in \mathbb{R}^{D \times D}.
\end{align}

Here, $\boldsymbol{\Sigma}_l$ is a diagonal matrix of singular values sorted in descending order. The top singular vector $\boldsymbol{v}_l^{\text{edit}}$ (i.e., the first column of $\boldsymbol{V}_l$) captures the dominant direction that separates positive from negative features. We use this vector as the steering direction, representing a latent shift that guides the model to focus more effectively on salient objects in the image.

During inference, we apply the learned steering vectors to edit the hidden representations of test samples. Specifically, for each layer $l$, we update the hidden feature $\vx_l$ as $\vx_l \leftarrow \vx_l + \boldsymbol{v}_l^{\text{edit}}$.

\subsection{Why \nameshort{} works?}
\label{how work}

We provide an analysis to understand what the model learns through the latent steering procedure. This analysis can be conducted for each transformer layer $l$; for simplicity, we drop the subscript $l$ and analyze layers independently. Let $f(\boldsymbol{x})$ denote the output of the LVLM given input features $\boldsymbol{x}$, and let $\mathbf{A}$ represent the attention matrix influenced by $\boldsymbol{x}$, $\mathbf{A}(\boldsymbol{x})$. For simplicity, we assume a single attention head. To approximate the model’s behavior under perturbed inputs, we apply a first-order Taylor expansion to estimate the output for a noise input $\tilde{\boldsymbol{x}}$, which is expressed as:
\begin{align}  
  f(\tilde{\mathbf{A}}) = f(\mathbf{A}) + (\frac{\partial f}{\partial \mathbf{A}})^\top (\tilde{\mathbf{A}} - \mathbf{A}) + \mathcal{R}
  ~\Leftrightarrow~ (\frac{\partial f}{\partial \mathbf{A}})^\top \mathbf{A} =  \mathbf{1}^\top ( {\color{blue}{\frac{\partial f}{\partial \mathbf{A}} \odot \mathbf{A}}})  = {\color{red}f(\mathbf{A})-f(\tilde{\mathbf{A}})},
  \label{diss}
\end{align}
where we suppose all matrices are vectorized and use $\mathbf{A}$ to denote $\mathbf{A}(\boldsymbol{x})$, and $\mathbf{1}$ and $\mathcal{R}$ are the all-one vector and 
higher-order infinitesimal term, respectively. Since $\tilde{\boldsymbol{x}}$ is assumed to be ideal noise, where tokens are independent of each other, the resulting attention matrix satisfies $\mathbf{A}(\tilde{\boldsymbol{x}}) = \mathbf{0}$. 

The \textcolor{blue}{blue} components in Eq.~\ref{diss} share the same formulation as the visual contribution maps computed by \nameshort{} in Eq.~\ref{eq:modified_att}. Additionally, we observe that the \textcolor{red}{red} term in Eq.\ref{diss} closely resembles recent decoding strategies for OH mitigation, such as VCD~\cite{leng2024mitigating} ($(1 + \alpha) f(\boldsymbol{x}) - \alpha f(\tilde{\boldsymbol{x}})$) debiasing the model’s prior-driven predictions. Based on this connection, we infer that applying the vision-aware masking via the visual contribution maps enables the resulting latent steering to eliminate model bias at the feature level, similar to the decoding-level as in VCD, and potentially mitigate OH.

\section{Experiments}
\label{sec:exp}
This section first evaluates the proposed \nameshort{} in OH mitigation tasks. Moreover, to gain insights into the hallucination behavior of modern open-source LVLMs, we conduct a series of visualization experiments and reveal several limitations of existing OH benchmarks. Finally, we further conduct an ablation and analysis experiment.

\noindent \textbf{Datasets.} We evaluate our model on a series of popular applied datasets for hallucination mitigation and general ability evaluation. More details about the datasets are provided in supplementary materials. For OH benchmark, we use CHAIR~\cite{rohrbach2018object}, AMBER~\cite{wang2023llm}, POPE~\cite{li2023evaluating}, MMHal~\cite{sun2024aligning} and MMVP~\cite{tong2022videomae} to test the performance of \nameshort{} in OH mitigation. Moreover, we implement Multi-modal Large Language Model Evaluation benchmark (MME)~\cite{fu2023mme}, Visual Reasoning and Compositional Question Answering (GQA)~\cite{hudson2019gqa} and LLaVA-Bench~\cite{liu2023improved} to test the general ability of the LVLMs.

\textbf{Implementation Details.} To evaluate the effectiveness of \nameshort{}, we implement \nameshort{} on three mainstream large vision-language models, including LLaVA-1.5~\cite{liu2024visual}, MiniGPT-4~\cite{zhu2023minigpt4} and Qwen2-VL~\cite{wang2024qwen2}. More details are provided in supplementary materials.

\subsection{OH Mitigation Results}

\begin{table*}[htp]
\centering
\setlength{\tabcolsep}{2pt}
\resizebox{\linewidth}{!}{
\begin{tabular}{
  l                              
  | c c c                        
  >{\centering\arraybackslash}p{1cm}
  >{\centering\arraybackslash}p{1.1cm}
  | c c c 
  >{\centering\arraybackslash}p{1cm} 
  >{\centering\arraybackslash}p{1.1cm} 
}
\toprule
\multirow{2}{*}{\textbf{Method}} 
&\multicolumn{5}{c|}{LLaVA-1.5}
&\multicolumn{5}{c}{MiniGPT-4} \\
\cmidrule{2-11}
&\textbf{C}$_S \downarrow$ 
&\textbf{C}$_I \downarrow$ 
&BLEU$\uparrow$ 
&F1 &Len 
&\textbf{C}$_S \downarrow$ 
&\textbf{C}$_I \downarrow$ 
&BLEU$\uparrow$ 
&F1 &Len \\ 

\midrule
Greedy
&$\text{20.4}_{\pm \text{2.8}}$ 
&$\text{7.1}_{\pm \text{0.3}}$ 
&$\text{15.7}_{\pm \text{0.1}}$ 
&\text{73.2}
&\text{54.7}
&$\text{32.4}_{\pm \text{2.2}}$ 
&$\text{12.2}_{\pm \text{0.4}}$ 
&$\text{14.6}_{\pm \text{0.1}}$ 
&\text{67.9}
&\text{55.4}
\\

Beam Search \cite{freitag2017beam}
&$\text{19.5}_{\pm \text{2.3}}$ 
&$\text{6.8}_{\pm \text{0.8}}$ 
&$\text{16.0}_{\pm \text{0.1}}$ 
&\text{71.7}
&\text{50.0}
&$\text{30.1}_{\pm \text{0.3}}$ 
&$\text{11.9}_{\pm \text{0.4}}$ 
&$\text{15.4}_{\pm \text{0.2}}$ 
&\text{67.4}
&\text{54.3}
\\

DoLa \cite{chuang2023dola}
&$\text{20.2}_{\pm \text{2.8}}$ 
&$\text{6.8}_{\pm \text{0.5}}$ 
&$\text{15.7}_{\pm \text{0.1}}$ 
&\text{72.5}
&\text{52.1}
&$\text{31.9}_{\pm \text{3.3}}$ 
&$\text{12.2}_{\pm \text{0.9}}$ 
&$\text{14.5}_{\pm \text{0.1}}$ 
&\text{68.1}
&\text{55.8}
\\

OPERA \cite{huang2024opera}
&$\text{17.5}_{\pm \text{0.5}}$ 
&$\text{6.1}_{\pm \text{0.3}}$ 
&$\text{16.0}_{\pm \text{0.1}}$ 
&\text{72.6}
&\text{53.1}
&$\text{29.7}_{\pm \text{0.3}}$ 
&$\text{12.0}_{\pm \text{0.3}}$ 
&$\text{14.8}_{\pm \text{0.1}}$ 
&\text{67.1}
&\text{54.6}
\\

VCD \cite{leng2024mitigating}
&$\text{20.3}_{\pm \text{1.1}}$ 
&$\text{7.3}_{\pm \text{0.1}}$ 
&$\text{14.5}_{\pm \text{0.0}}$ 
&\text{71.0}
&\text{51.6}
&$\text{29.0}_{\pm \text{2.8}}$ 
&$\text{12.6}_{\pm \text{1.2}}$ 
&$\text{14.4}_{\pm \text{0.0}}$ 
&\text{66.2}
&\text{53.1}
\\

Woodpecker \cite{yin2023woodpecker}
&$\text{23.9}_{\pm \text{4.6}}$ 
&$\text{7.5}_{\pm \text{0.1}}$ 
&$\text{17.1}_{\pm \text{0.0}}$ 
&\text{-}
&\text{-}
&$\text{28.9}_{\pm \text{2.2}}$ 
&$\text{10.2}_{\pm \text{0.9}}$ 
&$\text{15.3}_{\pm \text{0.0}}$ 
&\text{-}
&\text{-}
\\

LURE \cite{zhou2023analyzing}
&$\text{19.5}_{\pm \text{2.4}}$ 
&$\text{6.5}_{\pm \text{0.4}}$ 
&$\text{16.0}_{\pm \text{0.0}}$ 
&\text{-}
&\text{-}
&$\text{27.9}_{\pm \text{2.3}}$ 
&$\text{10.2}_{\pm \text{0.9}}$ 
&$\text{15.0}_{\pm \text{0.1}}$ 
&\text{-}
&\text{-}
\\

HALC \cite{chen2024halc}
&$\text{16.9}_{\pm \text{2.1}}$ 
&$\text{5.7}_{\pm \text{0.6}}$ 
&$\text{16.0}_{\pm \text{0.1}}$ 
&\text{71.2}
&\text{51.0}
&$\text{25.2}_{\pm \text{2.0}}$ 
&$\text{9.4}_{\pm \text{0.4}}$ 
&$\text{14.9}_{\pm \text{0.1}}$ 
&\text{67.4}
&\text{53.8}
\\

VTI-vision \cite{liu2025reducing}
&$\text{17.4}_{\pm \text{2.0}}$ 
&$\text{6.0}_{\pm \text{0.6}}$ 
&$\text{15.5}_{\pm \text{0.1}}$ 
&\text{73.3}
&\text{54.8}
&$\text{30.4}_{\pm \text{1.6}}$ 
&$\text{11.5}_{\pm \text{0.6}}$ 
&$\text{15.1}_{\pm \text{0.1}}$ 
&\text{67.4}
&\text{54.8}
\\

\midrule
\rowcolor{mygray} \textbf{\nameshort} 
&$\textbf{16.7}_{\pm \text{2.4}}$ 
&${\textbf{5.7}}_{\pm \text{0.8}}$ 
&$\text{15.8}_{\pm \text{0.1}}$ 
&\text{72.3}
&\text{54.5}
&$\text{27.7}_{\pm \text{1.7}}$ 
&$\text{11.2}_{\pm \text{0.8}}$ 
&$\text{15.0}_{\pm \text{0.1}}$ 
&$\text{67.6}$
&$\text{53.6}$
\\ 
\bottomrule
\end{tabular}
}
\vspace{-6pt}
\caption{CHAIR evaluation results on MSCOCO with different methods for mitigating OH. We use 64 as the max token number in this experiment.}
\label{tab:chair_results}
\vspace{-8pt}
\end{table*}

\paragraph{Compared to Existing OH Mitgation Methods.} Table~\ref{tab:chair_results} summarizes the performance of \nameshort{} when incorporated into LLaVA-1.5 and MiniGPT-4, in comparison with existing hallucination mitigation approaches. LLaVA enhanced with \nameshort{} outperforms all compared methods, while MiniGPT-4 combined with \nameshort{} achieves performance comparable to most decoding-based baselines. Among the metrics,C$_\text{S}$ is particularly critical, as a caption containing multiple correct objects but a single hallucinated one is still considered erroneous. A substantial improvement in C$_\text{S}$ indicates that \nameshort{} effectively eliminates the remaining hallucinated objects. We also report BLEU, F1, and Length (Len) metrics to ensure that \nameshort{} does not compromise response quality or object coverage. The results show almost no degradation, confirming the quality preservation of generated captions.

\begin{table*}[htp]
\centering
\setlength{\tabcolsep}{2pt}
\resizebox{\linewidth}{!}{
\begin{tabular}{
  l                              
  | p{1.1cm}<{\centering} p{1.1cm}<{\centering} p{1.1cm}<{\centering}                        
  | p{1.1cm}<{\centering} p{1.1cm}<{\centering} p{1.1cm}<{\centering}  p{1.1cm}<{\centering} | p{1.1cm}<{\centering} p{1.1cm}<{\centering}
  | p{1.1cm}<{\centering} p{1.1cm}<{\centering}
  | p{1.1cm}<{\centering} p{1.1cm}<{\centering} 
  | p{1.1cm}<{\centering} 
}
\toprule
\multirow{2}{*}{\textbf{Model}} 
&\multicolumn{3}{c|}{CHAIR}
&\multicolumn{6}{c|}{AMBER}
&\multicolumn{2}{c|}{POPE}
&\multicolumn{2}{c|}{MMHal}
&{MMVP} \\
\cmidrule{2-4} \cmidrule{5-10} \cmidrule{11-12} \cmidrule{13-15}
&\textbf{C}$_{S \downarrow}$ 
&\textbf{C}$_{I \downarrow}$ 
&F1 
&CH. $_\downarrow$
&Co. $_\uparrow$
&Hal. $_\downarrow$
&Cog. $_\downarrow$
&Acc. $_\uparrow$
&F1 $_\uparrow$
&Acc. &F1
&Score$_\uparrow$ &Hal.$_\downarrow$
& Score$_\uparrow$\\
\midrule
\text{LLaVA-1.5} & 48.7 & 13.4 & 77.0 & 7.2 & 50.6 & 32.5  & 3.7 & 71.9  & 74.8 & 81.4 & 79.7 & 2.6 & 60.4 & 26.7 \\
\rowcolor{mygray}\textbf{\nameshort} & \textbf{36.2} & \textbf{10.0} & 78.2 & \textbf{5.5} & \textbf{51.2} & \textbf{28.0} & \textbf{2.7} & \textbf{74.7}  & \textbf{78.5} & \textbf{82.7} & \textbf{84.1} & 2.7 & \textbf{56.3} & \textbf{31.3} \\
\midrule
\text{Qwen2-VL}   & 44.4 & 8.71 & 75.2 & \text{6.9} & \text{71.7} & \text{58.3}  & \text{6.1} & \text{78.6}  & \text{83.2} & 84.4 & 82.4 & 3.7 & 38.5 & 51.3 \\
\rowcolor{mygray}\textbf{\nameshort} & \textbf{39.6} & 8.66 & 75.3 & \text{6.3} & \text{70.3} & \textbf{49.1}  & \textbf{5.2} & \text{78.9}  & \text{84.0} & \textbf{86.3} & \textbf{85.8} & 3.9 & \textbf{32.3} & 52.7 \\
\bottomrule
\end{tabular}
}
\vspace{-5pt}
\caption{Evaluation results on the CHAIR~\cite{rohrbach2018object}, AMBER~\cite{wang2023llm}, POPE~\cite{li2023evaluating}, MMHal~\cite{sun2024aligning} and MMVP~\cite{tong2022videomae} datasets.}
\label{tab:amber_combined}
\vspace{-18pt}
\end{table*}

\paragraph{Results on Hallucination Benchmarks.}
We further evaluate the effectiveness of \nameshort{} in mitigating object hallucination (OH) by applying it to LLaVA-1.5 and Qwen2-VL across multiple benchmarks, including CHAIR (512 max-token setting), AMBER, POPE, MMHal, and MMVP, as presented in Table~\ref{tab:amber_combined}. The results show that integrating \nameshort{} consistently improves performance compared to the original models on most benchmarks. For CHAIR, the F1 scores remain comparable or even slightly higher than those of the original LVLMs, indicating that both object precision and recall are preserved. Notably, improvements on Qwen2-VL are more moderate compared to LLaVA-1.5. This may be attributed to the multi-scale vision encoder and complex visual features in Qwen2-VL, which make it more difficult to trace the influence of visual tokens on output tokens, thereby reducing the effectiveness of latent steering. On the POPE benchmark, both models show clear improvements with \nameshort{}. For MMHal-Bench, although the overall average score improvements are modest, \nameshort{} significantly reduces hallucination rates. Specifically, LLaVA-1.5’s hallucination rate drops from 60.4 to 56.3, and Qwen2-VL’s rate decreases markedly from 38.5 to 32.3. In contrast, \nameshort{} shows limited improvement on the MMVP benchmark. which may be due to the multiple-choice questions format of MMVP tasks.

\begin{figure}[htp]
  \centering
  \begin{minipage}[b!]{0.5\textwidth}
    \centering
    \includegraphics[width=\linewidth]{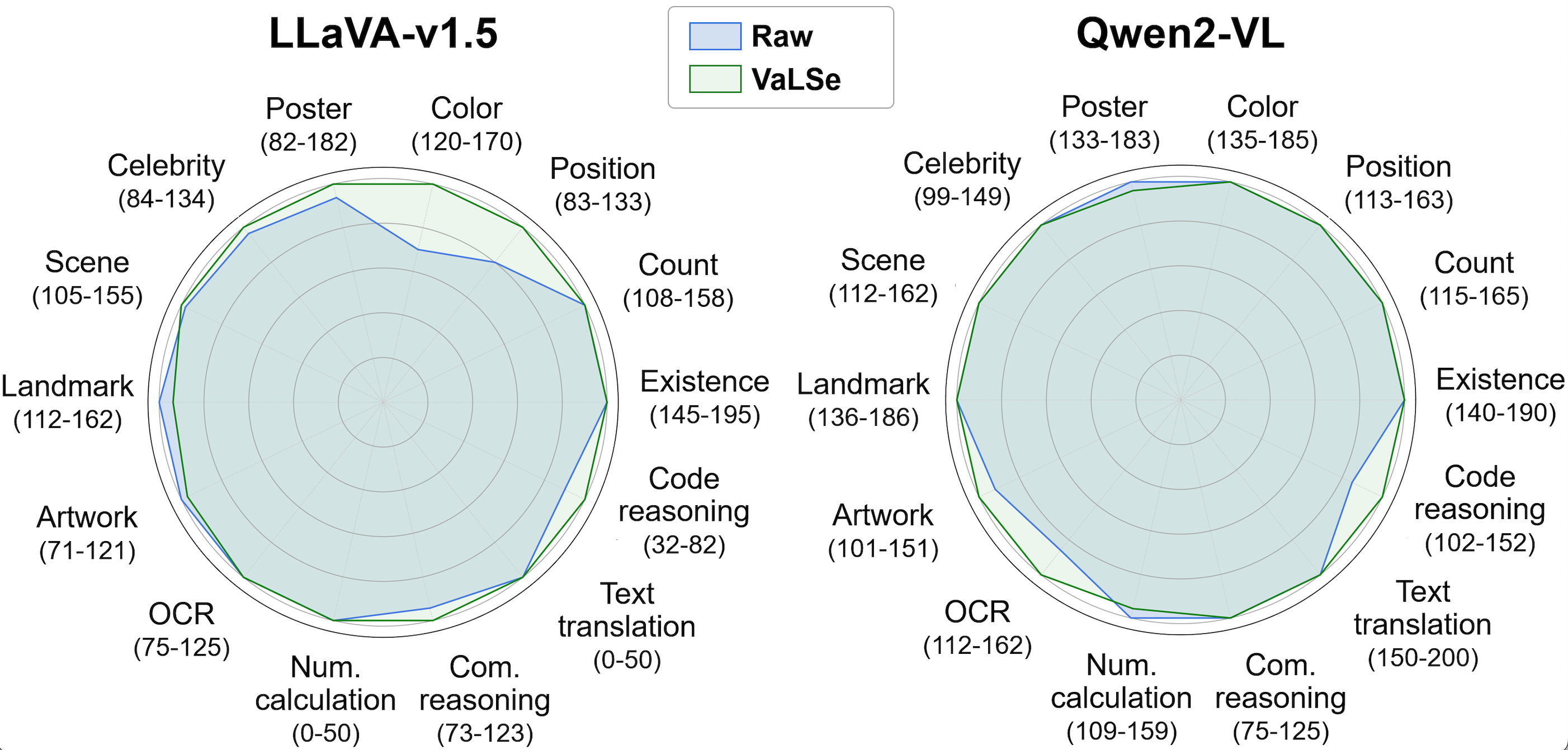} 
    \caption{Results on MME.}
    \label{fig-mme}
  \end{minipage}%
  \hfill
  \begin{minipage}[b!]{0.49\textwidth}
    \vspace{20pt}
    \centering
    \resizebox{1.0\linewidth}{!}{
\begin{tabular}{
  l
  | c c c
  | c c
}
\toprule
\multirow{2}{*}{\textbf{Model}} 
& \multicolumn{3}{c|}{{GQA}} 
& \multicolumn{2}{c}{{LLaVA-Bench}} \\
\cmidrule{2-6}
& {Binary} & {Open} & {Acc.} 
& {Acc.} & {Detail.} \\
\midrule
\text{LLaVA-1.5} & 77.9 & 47.1 & 61.2 & 5.4 & 5.2 \\
\rowcolor{mygray}\textbf{\nameshort} & 78.3 & 46.9 & 61.3 & 6.2 & 5.8 \\
\midrule
\text{Qwen2-VL} & 83.1 & 45.1 & 62.5 & 7.0 & 6.5 \\
\rowcolor{mygray}\textbf{\nameshort} & 82.6 & 45.3 & 62.4 & 7.3 & 6.5 \\
\bottomrule
\end{tabular}
}
\captionof{table}{Results on GQA and LLaVA-Bench.}
\label{tab:multi_eval}
  \end{minipage}
\vspace{-10pt}
\end{figure}

\paragraph{General Task Performance.}
We evaluate both the original LVLMs and their \nameshort{}-enhanced versions on MME, GQA, and LLaVA-Bench to assess whether \nameshort{} impacts the models’ general capabilities. As shown in Figure~\ref{fig-mme}, LLaVA-1.5 exhibits improved performance in color and positional understanding, while Qwen2-VL shows notable gains in OCR and code-related tasks. Additionally, Table~\ref{tab:multi_eval} reports results on GQA and LLaVA-Bench, demonstrating that model performance remains comparable to, or even surpasses, that of the original baselines. These results suggest that \nameshort{} effectively mitigates object hallucination without compromising the general reasoning or multimodal capabilities of the underlying LVLMs.

\subsection{Are these OHs Indeed Hallucinated Objects?}
\vspace{-5pt}
\begin{figure}[t!]
  \centering
  \includegraphics[width=\linewidth]{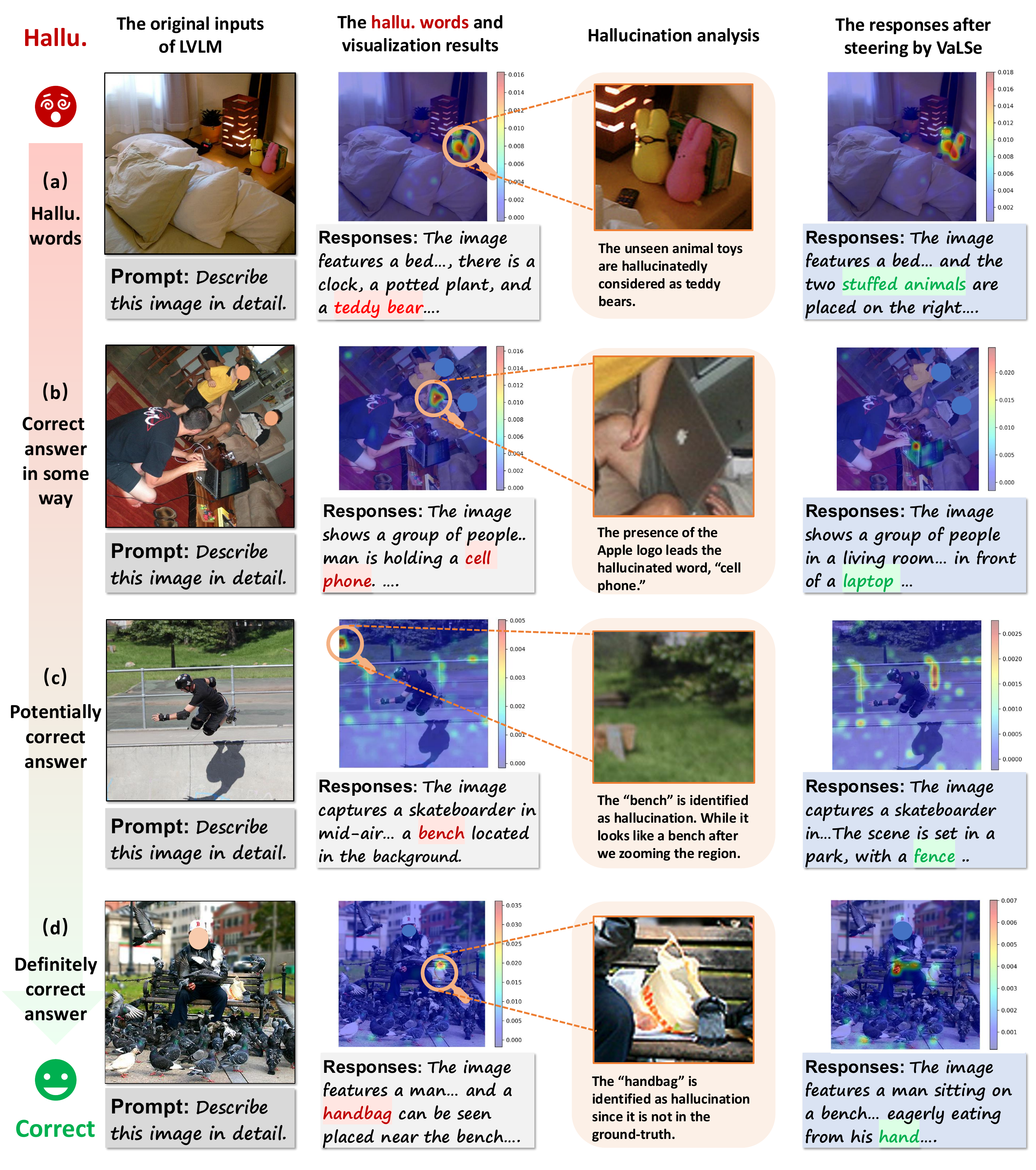}
  \vspace{-20pt}
  \caption{The visualization and analysis results via \nameshort{} of four different types of hallucination using LLaVA-1.5 on the CHAIR benchmark.}
  \label{fig-vis}
\vspace{-20pt}
\end{figure}
In our experiments, we systematically analyze the hallucinated words generated by LLaVA as identified by the CHAIR metric, with results shown in Figure~\ref{fig-vis}. The figure is organized into four columns: (1) the original input image, (2) the hallucinated word along with its visual contribution map generated by \nameshort{}, (3) a zoomed-in crop of the original image corresponding to high-activation regions in the contribution map, and (4) the response generated by the model after being edited by \nameshort{} using the visualization result. From these examples, we identify and categorize four distinct types of hallucination present in the CHAIR benchmark.

\textbf{Truly Hallucinated Words.} Figure~\ref{fig-vis} (a) presents a typical case of object hallucination, where the model incorrectly identifies unseen animal toys as teddy bears. This hallucinated prediction is effectively corrected by \nameshort{}, which steers the model’s attention more focus on the visual cues.

\textbf{Factual Hallucinated Words.} Figure~\ref{fig-vis}~(b) illustrates a more interesting example. Here, the model makes a factual hallucination, describing the presence of a cell phone due to the appearance of an Apple logo in the image. While the logo is on a laptop and no phone is present, the hallucination reflects a strong prior association within the LVLM, linking the Apple logo with the cell phone concept. However, such a prediction could be viewed as reasonable in some way. After all, given an Apple logo, the first word that comes to our mind can be ``iPhone'', corresponding to a cell phone.

\textbf{Unclear Hallucinated Words.} CHAIR may also flag potentially correct answers as hallucinations. As shown in the zoomed-in region of Figure~\ref{fig-vis}~(c), there appears to be a vague object resembling a bench on the grass. However, due to its small size and ambiguous appearance, it is difficult to definitively determine whether the word bench constitutes a hallucination. 

\textbf{False Hallucinated Words.} Last but most importantly, Figure~\ref{fig-vis}~(d) presents a case where the CHAIR metric flags a word as hallucinated, despite it being a correct prediction. In this example, the model accurately identifies a handbag in the image. However, because the handbag is not a prominent object, it is not included in the ground-truth annotations, leading CHAIR to incorrectly consider it as a hallucination. This case highlights a key limitation of CHAIR: its reliance on incomplete or overly strict ground-truth labels, which can be a main limitation for CHAIR.

Despite the limitations of CHAIR, \nameshort{} still mitigates OH across all four identified types of hallucination. By applying vision-aware latent steering, \nameshort{} guides LLaVA to focus more on the main objects within the image, while avoiding unnecessary descriptions of ambiguous or visually uncertain regions. As a result, we observe a consistent reduction in both \textbf{C}$_{S}$ and \textbf{C}$_{I}$.

\vspace{-5pt}
\subsection{Ablation studies and further analysis}
\begin{figure}[htp]
\vspace{-10pt}
  \centering
  \includegraphics[width=\linewidth]{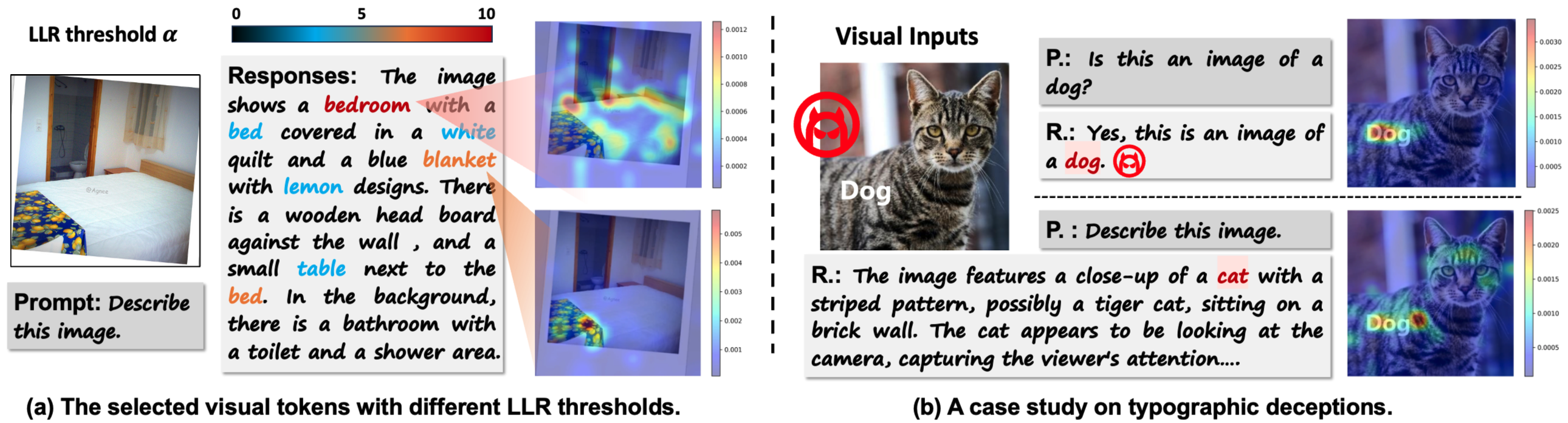}
  \vspace{-10pt}
  \caption{Further analysis with visualization results using LLaVA-1.5.}
  \label{figllr}
\vspace{-10pt}
\end{figure}

\vspace{-5pt}
\textbf{Selected Visual Tokens.} We present an analytical study to examine which types of words are identified as visual-based tokens, and how the selection threshold for LLR $\alpha$ influences the selection process. The results are shown in Figure~\ref{figllr}~(a). As expected, decreasing $\alpha$ results in more tokens being selected as visual-based. Furthermore, we observe that object-related words and attribute-related words, such as those describing color, are more likely to be selected, which meets our intuition.

\textbf{A Case Study for Wider Applications of \nameshort{}.} The example in Figure~\ref{figllr}~(b) provides a case study demonstrating how \nameshort{} can serve as an interpretability tool for analyzing typographic deception attacks~\cite{avrahami2022blended,cheng2024}. The results show that when the attack is successful, the model’s attention is misdirected by the word ``Dog''. However, when prompted to describe the image directly, the LVLM focuses on the stripe and the cat’s face, and produces the correct answer, even though it still exhibits high attention on the deceptive word ``Dog''. This case highlights that \nameshort{} is not only effective for mitigating OH, but also generalizes to broader interpretability tasks for modern LVLMs.

\textbf{Additional Results.} We provide a demo of \nameshort{}, the evaluation of visualization ability, and more ablation studies in supplementary materials due to the limitation of pages.

\vspace{-5pt}
\section{Conclusion}
\vspace{-5pt}
In this paper, we introduced \nameshort{}, which follows an interpretation-then-mitigation strategy, leveraging visual contribution maps to trace how visual inputs influence token-level outputs, and performing vision-aware latent space steering to enhance the model’s focus on vision-aware contents and reduce OH. Our experiments demonstrate that \nameshort{} achieves superior OH mitigation performance while maintaining general ability. Additionally, we highlight essential limitations in current OH benchmarks that can identify false hallucinations during evaluation. These findings suggest a more comprehensive evaluation benchmark for OH and that interpretability should play a more critical role in future research on hallucination mitigation. We also hope \nameshort{} provides both a practical solution and an analytical lens for advancing reliable and transparent LVLM systems.

{
    \small
    \bibliographystyle{unsrt}
    \bibliography{main_oh}
}





\newpage

\appendix

\section{Broader Impacts}
This work presents \nameshort{}, a training-free framework for mitigating object hallucination in LVLMs while enhancing interpretability through visual contribution mapping. As LVLMs are increasingly used in safety-critical domains, improving their transparency and reliability has broad societal value. \nameshort{} enables users to trace model decisions and identify hallucinated or biased outputs, supporting more trustworthy deployment. Providing an effective way to interpret how a specific output token is generated based on the visual inputs, it also exposes limitations in current evaluation benchmarks, motivating the development of more nuanced metrics. Moreover, by revealing model attention patterns, \nameshort{} could also inform adversarial strategies, underscoring the need to balance interpretability with security. Overall, this work contributes toward more robust, transparent, and responsible multimodal AI systems built based on LVLMs.

\section{Limitations and future works} 
While \nameshort{} provides an effective, training-free approach for mitigating object hallucination and interpreting visual-token interactions, it still has several limitations.

The quality of the visualization results heavily depends on how visual features from the encoder are integrated into the language model. In LVLMs such as LLaVA~\cite{liu2023improved} and LLaVA-Phi~\cite{zhu2024llava}, visual features are directly aligned with the language model via modules (such as linear layers) that preserve the spatial structure of the original visual inputs, allowing \nameshort{} to effectively trace how visual inputs influence text token generation. In contrast, models like MiniGPT-4~\cite{zhu2023minigpt4} and Qwen2-VL~\cite{wang2024qwen2} employ a Q-former to compress and blend visual features, followed by operations such as pixel-shuffle~\cite{shi2016real} to reduce the number of visual tokens. These transformations can destroy the original spatial relationships among tokens, degrading the quality of the contribution maps generated by \nameshort{}. Moreover, Qwen2-VL~\cite{wang2024qwen2} further employs the multi-scale visual feature extraction in the vision encoder, making it more difficult to interpret the generated visual contribution maps.
\vspace{-5pt}
\begin{figure}[htp]
  \centering
  \includegraphics[width=\linewidth]{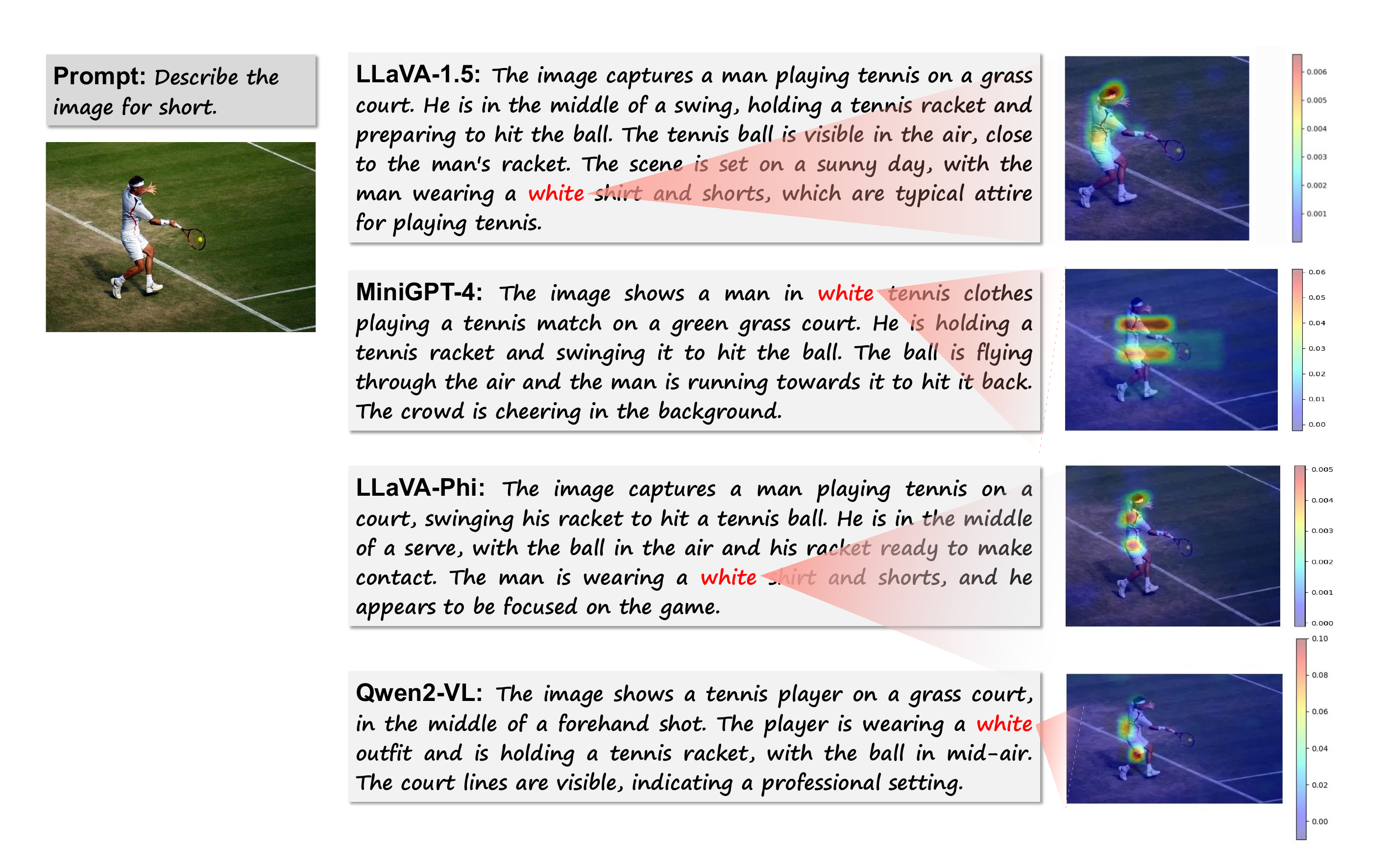}
  \caption{Visualization results of different four LVLMs using \nameshort{}.}
  \label{fig:models}
\end{figure}

We further infer that the conclusion in~\cite{neo2024towards}—which suggests that object information is highly localized to token positions corresponding to their original spatial location in the image—may only hold for models such as LLaVA and LLaVA-Phi. This is consistent with the authors’ discussion of limitations in their study. Moreover, we align with findings from prior work~\cite{xing2025large}, highlighting that many recent LVLMs adopt multi-resolution or multi-encoder architectures, complicating the alignment of intermediate features with their original spatial regions. These design choices pose inherent challenges for interpretability methods that rely on token-level spatial correspondence.

Figure~\ref{fig:models} provides a qualitative comparison across different LVLM architectures, illustrating notable differences in visualization clarity. This may explain why the effectiveness of latent steering varies across models: improvements on Qwen2-VL and MiniGPT-4 are less pronounced than those observed on LLaVA-1.5, likely due to reduced interpretability and weaker steering signals derived from less spatially coherent features. Nevertheless, applying \nameshort{} to systematically study OH in LLaVA yields several valuable insights into the limitations of current benchmark evaluations. These findings underscore the need for more nuanced, visually grounded assessment methods—an important direction for future work.

\section{Datasets}
\subsection{Datasets for Hallucination Evaluation}
\paragraph{CHAIR} CHAIR~\cite{rohrbach2018object} introduces a set of caption-image relevance metrics designed to evaluate the occurrence of object hallucinations (OH). This tool assesses image descriptions by comparing them with reference captions from standard datasets such as MSCOCO. The metrics quantify hallucinations based on the proportion of mentioned objects that are absent from the ground-truth object set, which is extracted from the reference captions.

Specifically, CHAIR$_S$ measures the proportion of generated captions that contain at least one hallucinated object, while CHAIR$_I$ quantifies the proportion of hallucinated objects among all generated objects. Lower scores indicate fewer hallucinations.
In our experiments, we also report BLEU to assess the overall quality of the generated text, and F1 score to evaluate the precision and recall of the generated objects relative to the ground-truth object set. For implementation, we randomly select 500 images from the MSCOCO 2014 validation set, repeating the evaluation three times. All methods are prompted with: “\textit{Please describe this image in detail.}”


\paragraph{AMBER} 
AMBER~\cite{wang2023llm} proposes an LLM-free, multi-dimensional benchmark consisting of 1,004 images. It includes both generative and discriminative tasks, providing a comprehensive evaluation of object hallucination. Specifically, the dataset contains 1,004 generation prompts and 14,216 discriminative prompts, which cover existence, attribute, and relation-based queries.

For evaluation, the generative task reports \textit{CHAIR} and \textit{Hal} scores to assess hallucinations in captions and object proportion. The \textit{Cover} metric measures the proportion of ground-truth objects included in the generated outputs, while \textit{Cog} evaluates the cognitive similarity between generated and target hallucinated objects—lower Cog scores indicate that hallucinated objects are easier to distinguish from real ones. The discriminative task reports accuracy and F1 score.

\paragraph{POPE} POPE~\cite{li2023evaluating} is a polling-based query framework for evaluating OH. It formulates a discriminative task by directly asking an LVLM whether a specific object is present in an image.

For implementation, each evaluation run samples 500 images from MSCOCO 2014 validation set. The method first extracts a set of candidate objects based on the segmentation results of the selected images. It then generates polling prompts in the form of \textit{“Is there a/an \{\} in the image?”}, where \{\} is filled with sampled object names using various strategies (random, popular, and adversarial). The evaluation focuses on the accuracy and F1 score of the model's responses, computed based on the statistical results of its positive and negative answers to the prompts.

\paragraph{MMHal-Bench} 
MMHal-Bench~\cite{sun2024aligning} is designed to evaluate response hallucinations in realistic user–LVLM interactions. The benchmark consists of 96 image-question pairs, where all questions are open-ended and span 8 question categories across 12 object-centric topics.

To assess hallucinations, GPT-4\cite{openai2023gpt4} is employed to analyze and rate LVLM responses. Each evaluation instance consists of the question, the corresponding model-generated response, the image category, and a standard human-written answer. These elements are incorporated into the prompt to support more accurate evaluation.

\paragraph{MMVP}
The MMVP benchmark~\cite{tong2024eyes} contains 150 multiple-choice questions and 300 images, where each question is associated with a pair of images. These image pairs constitute CLIP-Blind sets—constructed based on high similarity in CLIP embeddings but with clear visual differences. The dataset is designed to evaluate hallucinations potentially arise from such visual representation ambiguities.

\subsection{Datasets for General performance Evaluation}
\paragraph{MME} MME~\cite{fu2023mme} is a comprehensive benchmark consisting of 14 sub-tasks designed to evaluate the perception and cognition abilities of LVLMs. Each sub-task has a full score of 200. For each image, two manually constructed questions are provided, and the utility score for each sub-task is determined by accuracy, calculated based on the correctness of individual question responses. In our experiments, we evaluated model performance across the full set of tasks.

\paragraph{GQA} GQA~\cite{hudson2019gqa} is a large-scale benchmark designed for real-world visual reasoning and compositional question answering. In our experiments, we use the \textit{test-dev-balanced} split for evaluation, which includes both binary and open-ended question types.

\paragraph{LLaVA-Bench} LLaVA-Bench (In-the-Wild)~\cite{liu2024visual} is a benchmark comprising 24 images from diverse real-world sources and 60 corresponding questions. Each image is accompanied by a detailed, manually written description. This dataset is used to assess the ability of LVLMs to handle challenging and open-ended tasks. Following~\cite{leng2024mitigating}, we leverage LLaVA-Bench for qualitative evaluation using GPT-4V-aided assessment.

\section{Experiment Settings}
\subsection{Models} We apply \nameshort{} to four representative LVLMs: LLaVA-v1.5-7b\footnote{\url{https://huggingface.co/liuhaotian/llava-v1.5-7b}}, Qwen2-VL-7B-Instruct\footnote{\url{https://huggingface.co/Qwen/Qwen2-VL-7B-Instruct}}, MiniGPT4-llama2-7b\footnote{\url{https://github.com/Vision-CAIR/MiniGPT-4}}, and Mipha-3B\footnote{\url{https://github.com/xmoanvaf/llava-phi}}. The model weights are obtained from official repositories on GitHub or Hugging Face. All experiments involving LLaVA-1.5 are conducted on NVIDIA RTX 4090 GPUs.

\subsection{Implementation Details of LVLMs} 

\paragraph{Paired Samples Construction.} To generate visual token contribution maps for visual-based tokens, we randomly select 200 images from the MSCOCO 2017 training set, following the image set provided in the GitHub repository of \cite{neo2024towards}. Each image is paired with its corresponding response generated by an LVLM, which serves as the negative sample. To ensure the responses focus primarily on the main objects within the scene, we use the prompt “\textit{Describe the image for short.}” and constrain the maximum output length to 64 tokens.

The construction of negative samples is guided by visual token selection and corresponding visualizations, which are controlled by the LLR threshold $\alpha$ and the masking ratio $p$. In our experiments, we set $\alpha$ to 1.8 for MiniGPT-4, and 3 for both LLaVA-1.5 and Qwen2-VL, with a masking ratio of $p = 0.9$ across all three models. All threshold values are empirically tuned to reduce the inclusion of words that are irrelevant to object content, based on the global LLR distribution.

Instead of masking a fixed percentage $p$ of tokens, we adopt an adaptive strategy by masking all tokens whose relevance scores are below the mean relevance value in the token-wise relevance map.

\paragraph{Intervention Strength on the Shift Direction.} Following VTI~\cite{liu2025reducing}, we intervene in the decoder of the LLM by shifting its latent states along the direction $\boldsymbol{v}_l^{\text{edit}}$. When extracting features at the MLP layer for paired samples, we use the propagated feature of the last token. The intervention strength, denoted by $\beta$, is set as follows: 0.4 for MiniGPT-4; for LLaVA-1.5, 0.5 on CHAIR and AMBER, and 0.4 on other experiments; for Qwen2-VL, 0.2 on MMVP and MME, and 0.5 on other experiments.

\section{Ablation Studies}

\begin{table}[htbp]
  \centering
  \caption{Impact of different $\alpha$ thresholds for selecting visual-based tokens on performance}
    \begin{tabular}{l|ccc|ccc}
    \toprule
    \multirow{2}[4]{*}{$\alpha$} & \multicolumn{3}{c|}{max=64} & \multicolumn{3}{c}{max=512} \\
\cmidrule{2-7}          
& \textbf{C}$_{S \downarrow}$ 
& \textbf{C}$_{I \downarrow}$   
& F1         
& \textbf{C}$_{S \downarrow}$ 
& \textbf{C}$_{I \downarrow}$   
& F1         \\
\cmidrule{1-7}    raw   & 20.2  & 6.4   & 73.4  & 47.8  & 13.4  & 78.0  \\
    1     & 16.4  & 5.3   & 72.8  & 38.0   & 10.7  & 77.9 \\
    3     & 15.4  & 5.2   & 73.3  & 36.2  & 10.2  & 78.6 \\
    5     & 16.6  & 5.2   & 73.2  & 36.6  & 10.0  & 78.6 \\
    7     & 16.6  & 5.1   & 73.1  & 37.2  & 10.2  & 78.2 \\
    \bottomrule
    \end{tabular}%
  \label{tab:llr_threshold}%
  \vspace{-8pt}
\end{table}%

We conduct thorough ablation studies on key steps of the \nameshort{} framework. In all experiments, \nameshort{} is applied to the LLaVA-1.5 model and evaluated on the CHAIR task. For each experiment, we report C$_\text{S}$ and C$_\text{I}$ scores to assess hallucination, along with the F1 score to evaluate response quality. The configuration that consistently achieves lower C$_\text{S}$ and C$
_\text{I}$ scores while maintaining a competitive F1 score is selected as the final setting.

\paragraph{Threshold $\alpha$ for Selection of Visual-Based Tokens in Positive Sample Construction}  
Within our framework, we use an LLR-based criterion with threshold $\alpha$ to guide the selection of tokens for visualization. All the setting with a masking percentage $p=0.9$. The effect of varying the threshold $\alpha$ is presented in Table~\ref{tab:llr_threshold}.

\paragraph{Mask percentage $p$ in Positive Sample Construction} 
Since positive samples are constructed based on the combined relevancy map of individual tokens, we introduce a masking percentage $p$ to control the proportion of low-relevance tokens being masked. This effectively determines the extent to which tokens with stronger visual associations are retained. In all experiments, we fix the threshold at $\alpha = 3$. As shown in Table~\ref{tab:mask_precent}, the best performance is achieved when $p = 0.9$.

\begin{table}[htbp]
  \centering
  \caption{Impact of different masking ratios $p$ in positive samples on performance}
    \begin{tabular}{c|ccc|ccc}
    \toprule
    \multirow{2}[4]{*}{$p$} & \multicolumn{3}{c|}{max=64} & \multicolumn{3}{c}{max=512} \\
\cmidrule{2-7}          
& \textbf{C}$_{S \downarrow}$ 
& \textbf{C}$_{I \downarrow}$   
& F1         
& \textbf{C}$_{S \downarrow}$ 
& \textbf{C}$_{I \downarrow}$   
& F1         \\
    \midrule
    raw   & 20.2  & 6.4   & 73.4  & 47.8  & 13.4  & 78 \\
    0.95  & 16.2  & 5.0   & 73.2  & 37.0  & 10.0  & 78.4  \\
    0.9   & 15.4  & 5.2   & 73.3  & 36.2  & 10.2  & 78.6 \\
    0.8   & 16.6  & 5.5   & 73.1  & 37.6  & 10.8  & 78.1 \\
    0.7   & 17.2  & 5.6   & 72.9  & 39.2  & 11.1  & 78.3 \\
    \bottomrule
    \end{tabular}%
  \label{tab:mask_precent}%
  \vspace{-8pt}
\end{table}%

\paragraph{Type of Masking Method}
Given the selected $\alpha$ and $p$ values, we further investigate the impact of different masking strategies. The approaches evaluated include: Gaussian noise (mean 0, standard deviation 0.1), Gaussian blur (kernel size set to at least one-quarter of the image’s shorter side), zero replacement (replacing the masked region with zero), and mean replacement (filling the masked region with the mean value of the image tensor). As shown in Table~\ref{tab:mask_type}, mean replacement consistently achieves the best performance across both the 64-token and 512-token maximum output settings, offering the most effective balance between hallucination suppression and answer quality.

\begin{table}[!h]
  \centering
  \caption{Performance comparison of different replacement strategies for masked regions in the image component of positive samples.}
    \begin{tabular}{l|ccc|ccc}
    \toprule
    \multicolumn{1}{c|}{\multirow{2}[4]{*}{Mask Strategy}} & \multicolumn{3}{c|}{max=64} & \multicolumn{3}{c}{max=512} \\
\cmidrule{2-7}          
& \textbf{C}$_{S \downarrow}$ 
& \textbf{C}$_{I \downarrow}$   
& F1         
& \textbf{C}$_{S \downarrow}$ 
& \textbf{C}$_{I \downarrow}$   
& F1         \\
    \midrule
    raw   & 20.2  & 6.4   & 73.4  & 47.8  & 13.4  & 78.0  \\
    Gauss noise & 18.4  & 6.3   & 74.2  & 48.4  & 13.2  & 77.2  \\
    Gauss blur & 18.2  & 5.7   & 73.1  & 35.8  & 10.7  & 77.7  \\
    zero  & 18.2  & 5.8   & 73.4  & 40.6  & 11.0  & 78.0  \\
    mean  & 15.4  & 5.2   & 73.3  & 36.2  & 10.2  & 78.6 \\
    \bottomrule
    \end{tabular}%
  \label{tab:mask_type}%
  \vspace{-8pt}
\end{table}%

\paragraph{Effectiveness of the Positive Sample Method}
Following the VTI method \cite{liu2025reducing}, we adopt a steering approach. Instead of using contrastive responses as \cite{liu2025reducing}, we employ contrastive images to steer the LLM. To validate the necessity of relevance-guided masking, we compare against a random masking baseline, replicating the image contrast setup in VTI-vision. As shown in Table~\ref{tab:ramdom_relevancy}, across various masking percentage settings, relevance-guided masking consistently yields fewer object hallucinations, as evidenced by lower C$_\text{S}$ and C$_\text{I}$ scores. Moreover, at the optimal masking percentage for both methods, the relevance-guided approach achieves a higher F1 score, indicating superior overall performance in response generation.

\begin{table}[!h]
  \centering
  \caption{Results of random masking and relevancy-guided masking method}
    \begin{tabular}{c|ccc|ccc}
    \toprule
    \multirow{2}[4]{*}{$p$} 
    & \multicolumn{3}{c|}{Random} 
    & \multicolumn{3}{c}{Relevancy-Guided (Ours)} \\
\cmidrule{2-7}          
    & \textbf{C}$_{S \downarrow}$ 
    & \textbf{C}$_{I \downarrow}$  
    & F1
    & \textbf{C}$_{S \downarrow}$ 
    & \textbf{C}$_{I \downarrow}$  
    & F1       
    \\
    \midrule
    raw   & 20.2  & 6.4   & 73.4  & 20.2  & 6.4   & 73.4  \\
    0.95  & 17.4  & 5.4   & 72.7  & 16.2  & 5.0   & 73.2  \\
    0.9   & 17.2  & 5.3   & 72.7  & 15.4  & 5.2   & 73.3 \\
    0.8   & 18.2  & 5.7   & 73.2  & 16.6  & 5.5   & 73.1 \\
    0.7   & 18.8  & 5.8   & 73.5  & 17.2  & 5.6   & 72.9 \\
    \bottomrule
    \end{tabular}%
  \label{tab:ramdom_relevancy}%
  \vspace{-8pt}
\end{table}%

\section{Quantitative Results of Visualization}

\begin{figure}[h]
  \centering
  \includegraphics[width=\linewidth]{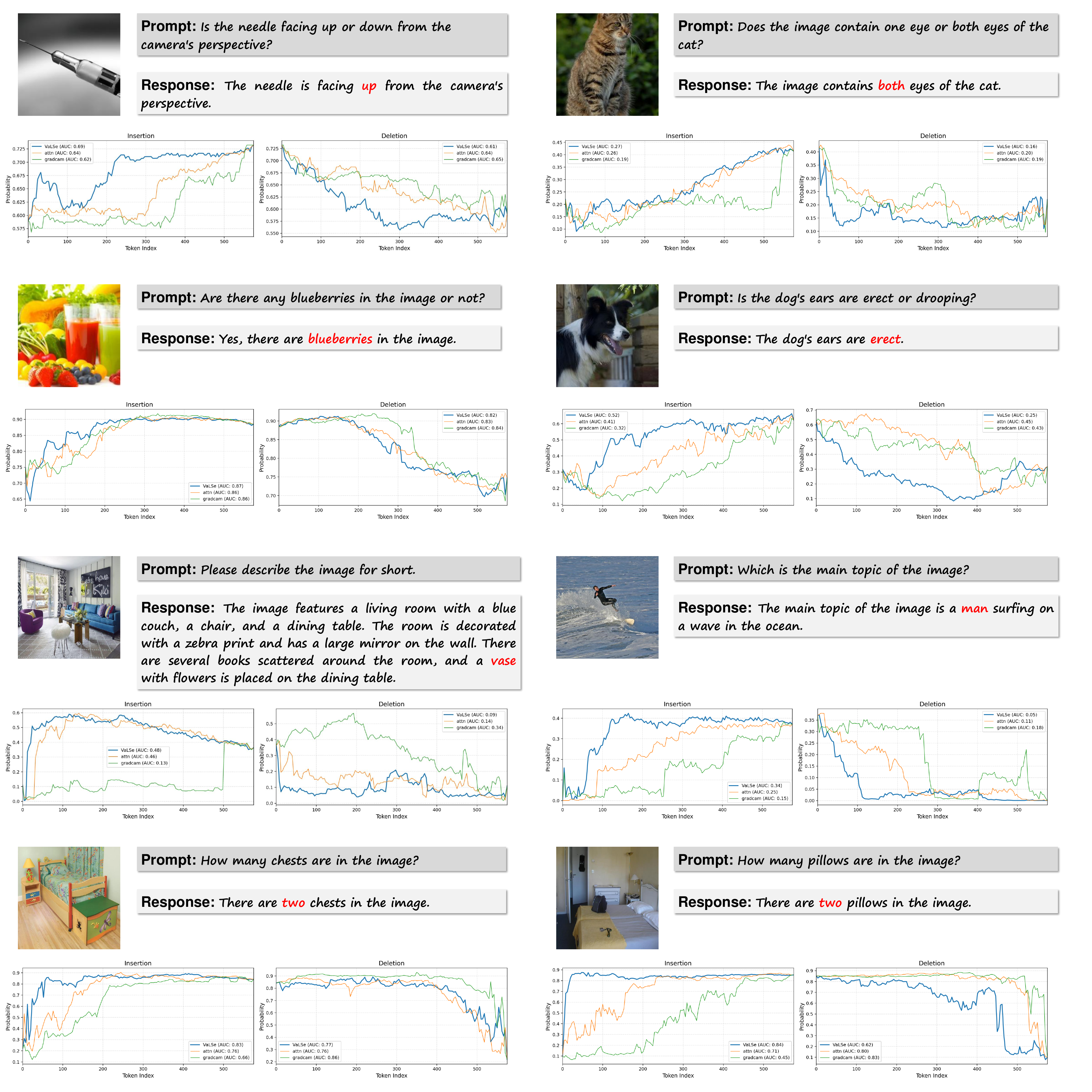}
  \caption{Insertion and deletion curves on 8 samples using three different visualization methods on LLaVA-1.5.}
  \label{fig:logits_ins_del}
\end{figure}

Following \cite{chefer2021generic}, we conduct deletion and insertion studies using LLaVA-1.5, comparing three visualization methods: attention maps, vision encoder Grad-CAM, and \nameshort{}, evaluated on 8 samples. For attention maps, we extract attention from the last layers (out of 32) of the LLM in LLaVA. The attention map is averaged across all heads within the selected layer.

For vision encoder Grad-CAM, we compute saliency maps with respect to the attention output after the layer normalization\footnote{Implementation based on \url{https://github.com/jacobgil/pytorch-grad-cam}} in the final layer, before features are passed into the LLM. We report and compare results from all three visualization methods. The outcomes are illustrated in Figure~\ref{fig:logits_ins_del}. The red words in the response correspond to the visualization tokens.

We briefly introduce the deletion and insertion experimental settings. Given visual inputs and text prompts, the LVLM generates a response. We then apply various visualization methods to produce visual contribution maps for a selected visual-based token. Ideally, if a contribution map accurately reflects the relevance between the token and visual content in the image, then masking the corresponding patch should significantly impact the token’s predicted probability.

In the insertion setting, we begin by masking the entire image with noise. Then, we gradually unmask patches one by one, ranked by their visual contribution scores. A better visualization method will reveal informative patches earlier, causing the token’s prediction probability to rise sooner in the process. 

In the deletion setting, we start with the original image and progressively mask patches in order of highest visual contribution. A better visualization method will remove important patches earlier, leading to a sharper drop in the token’s prediction probability early in the procedure.

As the results show, both \nameshort{} and the attention maps outperform Grad-CAM from the vision encoder in the insertion setting, achieving higher area under the curve (AUC) values and earlier rises in their respective curves. Notably, the curves do not exhibit a consistent trend when removing or inducing patches, primarily due to the presence of tokens preceding the visualization token, and possibly also due to the large number of parameters in the LLM. An opposite trend is observed in the deletion setting, where lower probability indicate that more relevant regions are being removed. 

Since \nameshort{} computes relevance maps by aggregating attention information across all layers, it achieves more stable and often better performance than a single-layer attention map. This demonstrates that \nameshort{} can effectively utilize internal attention signals in a model-agnostic manner.

\section{Gradio Demo for LVLM Visualization}

\begin{figure}[!h]
  \centering
  \includegraphics[width=0.6\linewidth]{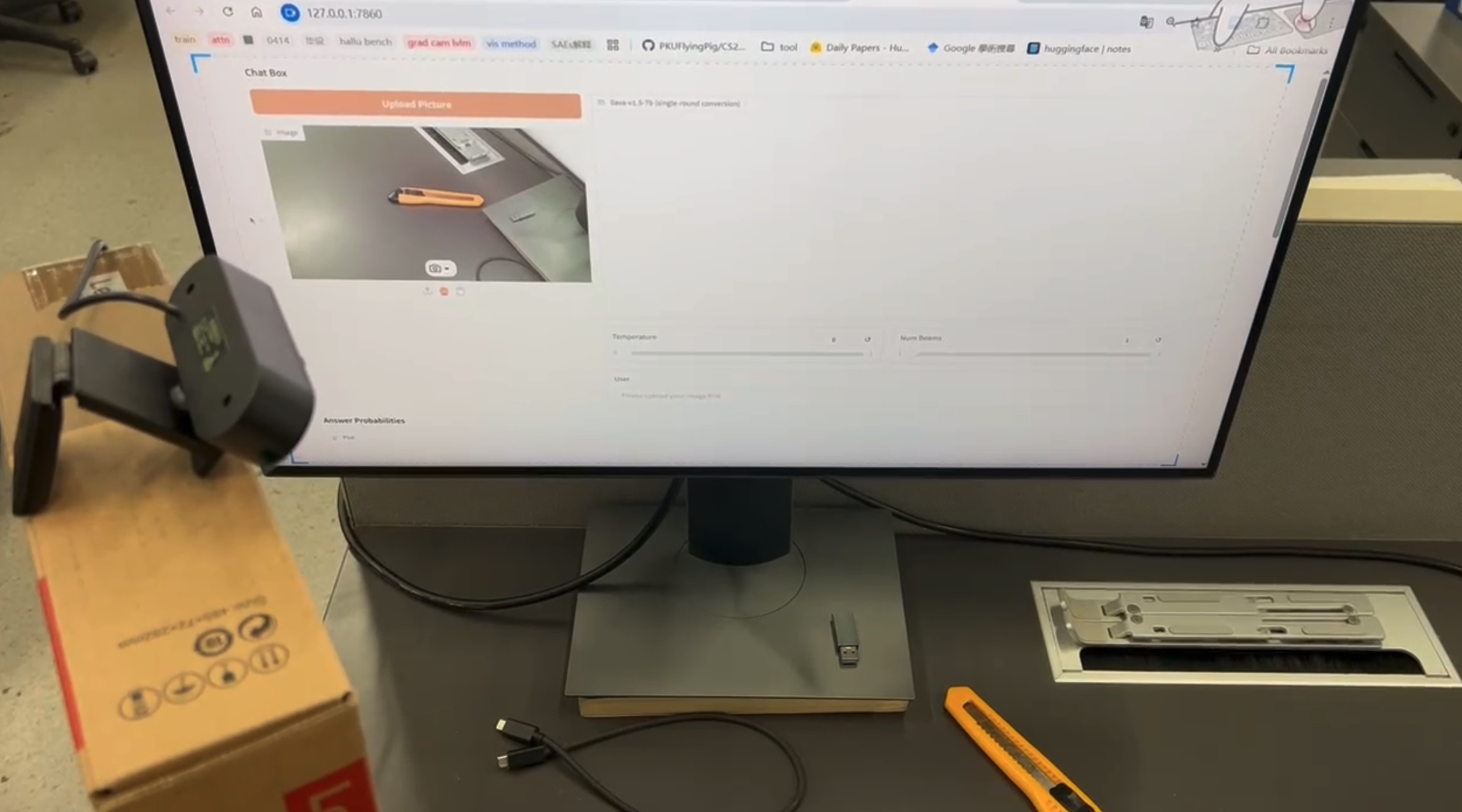}
  \caption{Real-world applications of the proposed system.}
  \label{fig:realworld}
\end{figure}

\begin{figure}[h]
  \centering
  \includegraphics[width=\linewidth]{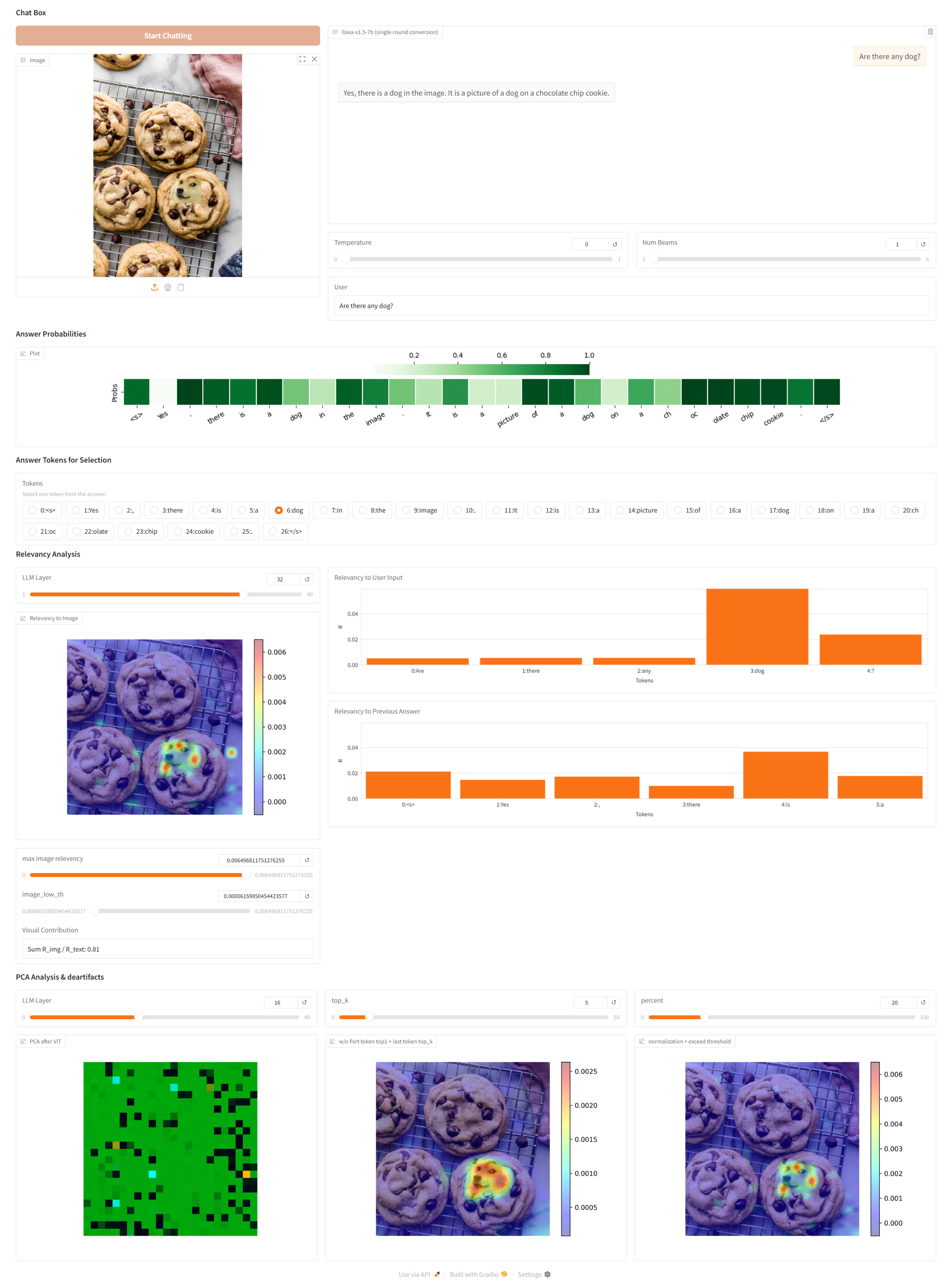}
  \caption{Gradio Demo of \nameshort{} for LLaVA-1.5-7b Visualization.}
  \label{fig:demo}
\end{figure}

To intuitively demonstrate our method, we develop an interactive Gradio\footnote{\url{https://www.gradio.app/}} demo for case studies, as illustrated in Figure~\ref{fig:demo}. The demo comprises three main components: a chatbot interface, a logits viewer, and a visualization module.

The visualization module is divided into two sections. The upper section presents raw results generated using the similar method to LVLM-Interpret~\cite{stan2024lvlmintrepret}, including LLM layer selection, visual relevance maps, and token-level text relevance scores. 

The left part of the lower section shows a PCA-based analysis of hidden states corresponding to image token indices across LLM layers. Empirically, in the middle-to-late layers, tokens with distinct orientations in the PCA space are indicative of potential artifacts.

On the right side, two de-artifacting strategies from \nameshort{} are provided. These methods aim to revise artifact-prone token regions by referencing non-semantic tokens (e.g., \texttt{<s>}, \texttt{<|endoftext|>}). The first method allows users to control the number of tokens to be replaced, while the second adjusts the replacement based on the cumulative relevance score ratio. To improve visual clarity when a large number of tokens are modified, a Gaussian filter is applied.

We also include a demonstration video \textbf{[Sample-1.mp4]} in the supplementary material to showcase the interface and its functionalities.

\paragraph{Real-world application.} With the Gradio, our visualization system can be deployed in real-world scenarios using a webcam. Figure~\ref{fig:realworld} shows an example captured in our lab. Using the webcam, we can perform visualization tests in open-world settings. A demonstration video \textbf{[Sample-2.mp4]} is also provided to showcase this setup.

\section{Additional Visualization Examples}
\begin{figure}[h]
  \centering
  \includegraphics[width=\linewidth]{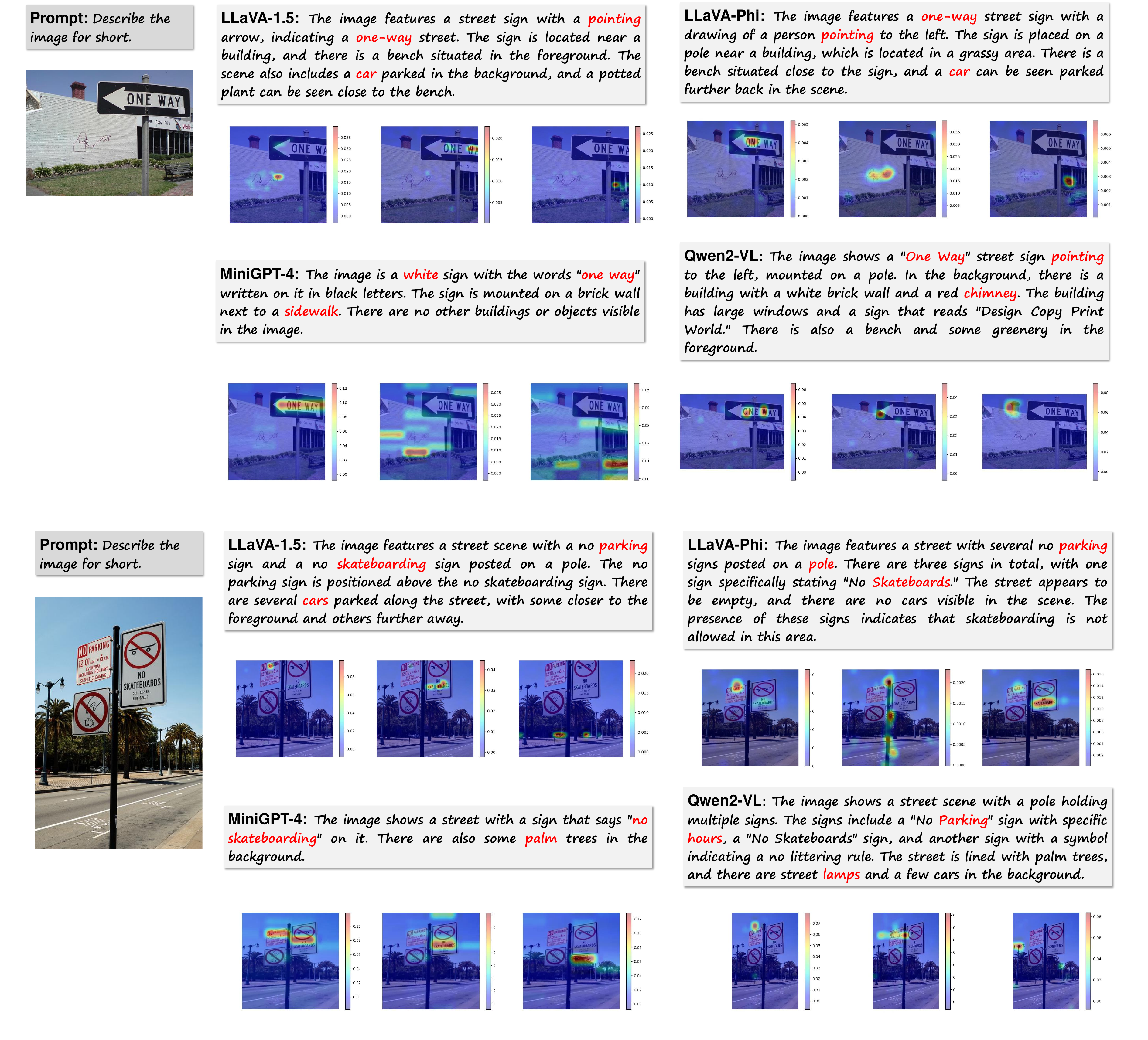}
  \caption{Additional visualization results for four different LVLMs using \nameshort{}.}
  \label{fig:more_vis1}
\end{figure}

\begin{figure}[h]
  \centering
  \includegraphics[width=\linewidth]{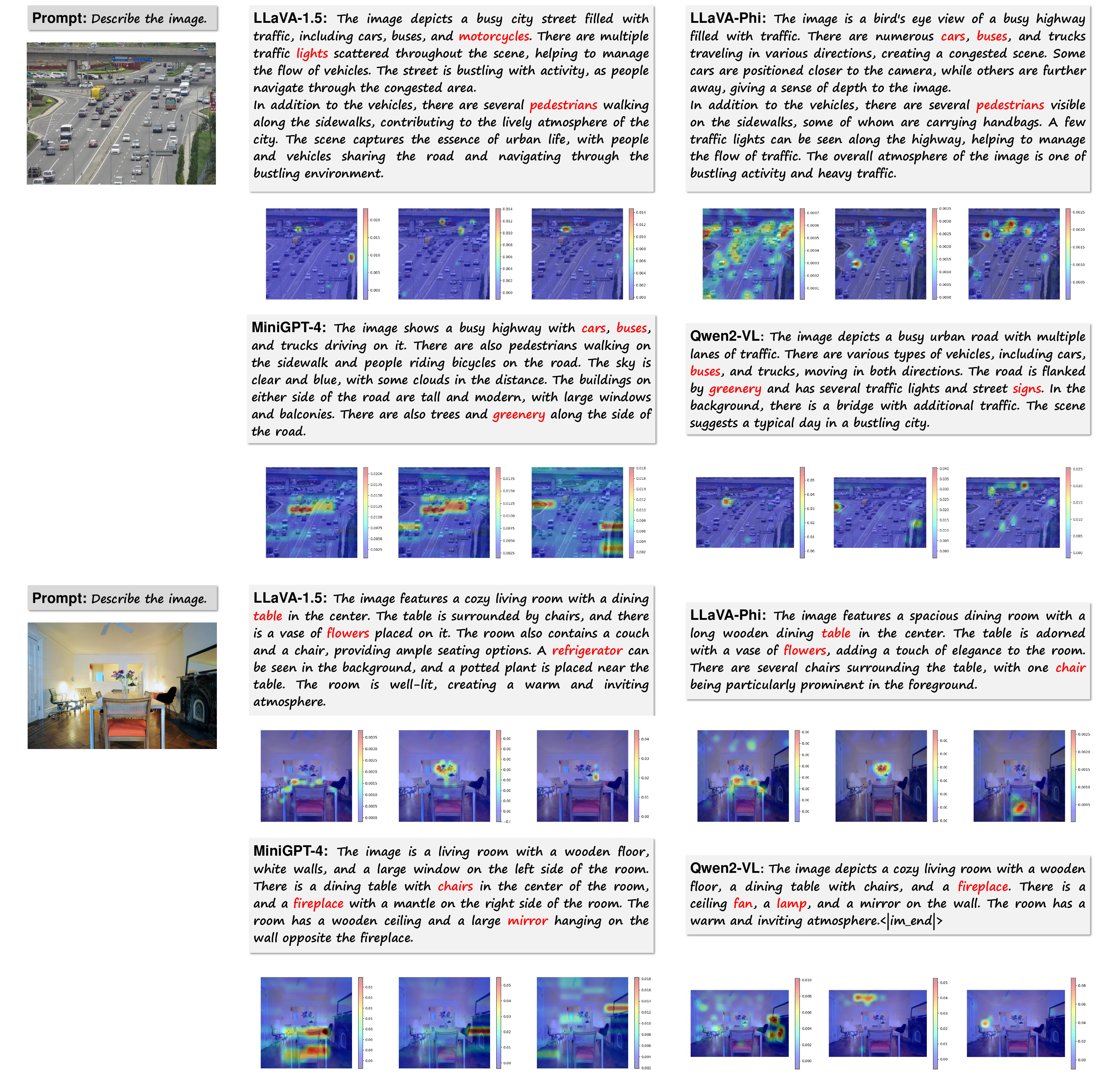}
  \caption{Additional visualization results for four different LVLMs using \nameshort{}.}
  \label{fig:more_vis2}
\end{figure}

We provide additional visualization examples for four LVLMs using \nameshort{}. As shown in Figure~\ref{fig:more_vis1} and Figure~\ref{fig:more_vis2}, each model response contains three highlighted words (in red). Visualizations corresponding to these words are presented in the images below the response, in the same order as the highlighted words.

\section{Cases of OH Mitigation}
\begin{figure}[h]
  \includegraphics[width=\linewidth]{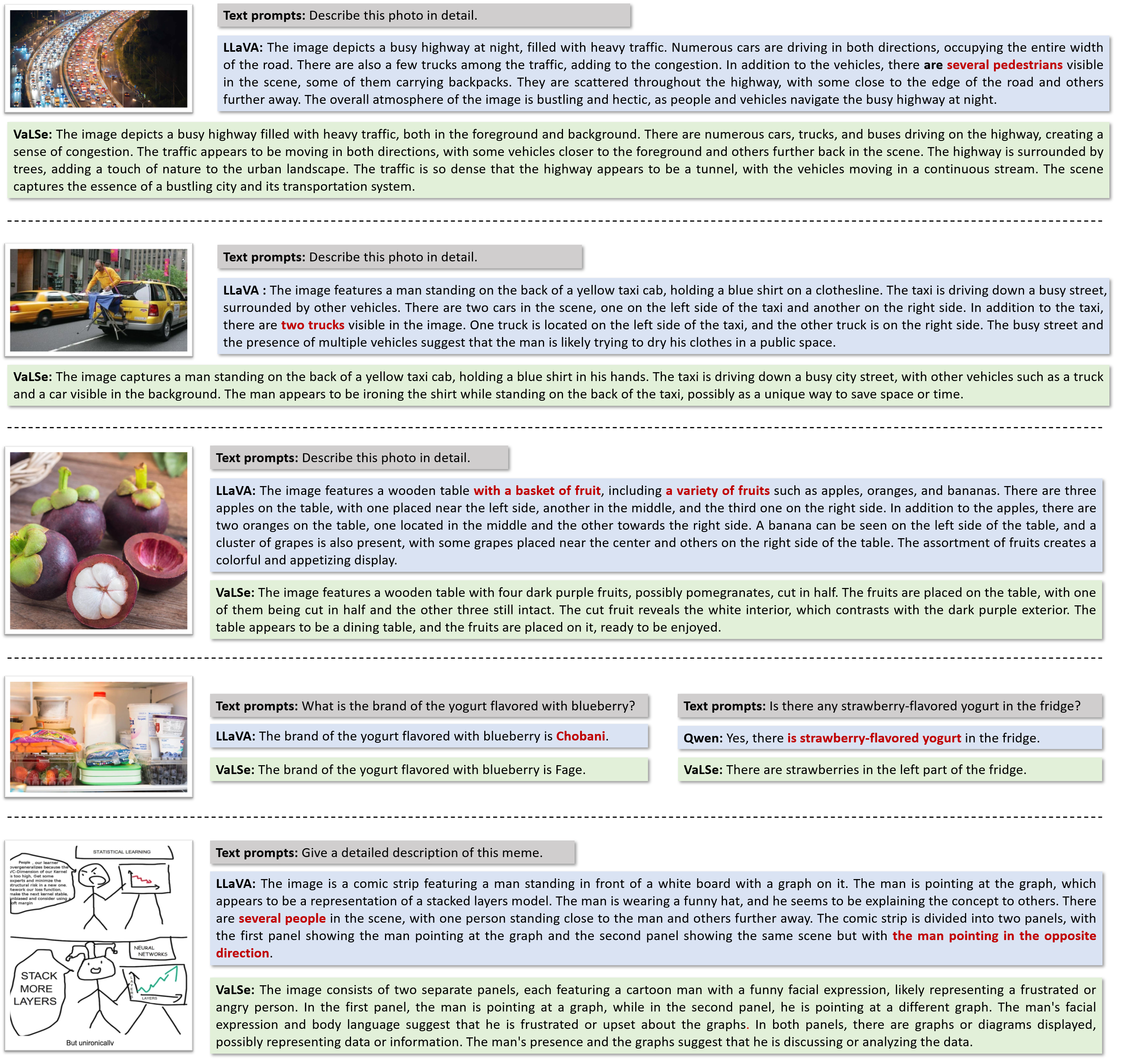}
  \caption{Cases of \nameshort{} on LLaVA-Bench.}
  \label{fig:llavabench_cases}
\end{figure}

Figure~\ref{fig:llavabench_cases} presents representative examples from LLaVA-Bench where \nameshort{} effectively mitigates object hallucinations (OH).

\end{document}